%% file: EA4PC.tex
\documentclass{article}

\PassOptionsToPackage{numbers}{natbib}

\usepackage[final]{neurips_2020}

\usepackage{caption}
\usepackage{subcaption}
\newcommand{\mcrot}[4]{\multicolumn{#1}{#2}{\rlap{\rotatebox{#3}{#4}~}}} 
\newcommand*{\twoelementtable}[3][l]%
{%
    \begin{tabular}[t]{@{}#1@{}}%
        #2\tabularnewline
        #3%
    \end{tabular}%
}

\usepackage[utf8]{inputenc} 
\usepackage[T1]{fontenc}    
\usepackage{hyperref}       
\usepackage{url}            
\usepackage{booktabs}       
\usepackage{amsfonts}       
\usepackage{nicefrac}       
\usepackage{microtype}      
\usepackage{xcolor}
\usepackage{shortcuts}
\usepackage{multicol}
\usepackage{wrapfig}
\usepackage{makecell}
\usepackage{cleveref}
\usepackage{multirow}
\Crefname{equation}{Eq.}{Eqs.}
\Crefname{figure}{Fig.}{Figs.}
\Crefname{tabular}{Tab.}{Tabs.}

\title{SE(3)-Transformers: 3D Roto-Translation Equivariant Attention Networks}

\makeatletter
\newcommand{\printfnsymbol}[1]{%
  \textsuperscript{\@fnsymbol{#1}}%
}
\makeatother

\author{
  Fabian B. Fuchs\thanks{equal contribution}$\,\,$\thanks{work done while at the Bosch Center for Artificial Intelligence} \\
  Bosch Center for Artificial Intelligence\\
  A2I Lab, Oxford University\\
  \texttt{fabian@robots.ox.ac.uk}
  \And
  Daniel E.~Worrall\printfnsymbol{1}\\
  Amsterdam Machine Learning Lab, Philips Lab\\
  University of Amsterdam\\
  \texttt{d.e.worrall@uva.nl} 
  \And
  Volker Fischer\\
  Bosch Center for Artificial Intelligence \\
  \texttt{volker.fischer@de.bosch.com}
  \And Max Welling \\
  $\quad\quad\,\,$ Amsterdam Machine Learning Lab $\quad\quad\,\,$ \\
  University of Amsterdam \\
  \texttt{m.welling@uva.nl}
}

\begin{document}

\maketitle

\begin{abstract}
We introduce the SE(3)-Transformer, a variant of the self-attention module for 3D point clouds and graphs, which is \emph{equivariant} under continuous 3D roto-translations. Equivariance is important to ensure stable and predictable performance in the presence of nuisance transformations of the data input. A positive corollary of equivariance is increased weight-tying within the model. The SE(3)-Transformer leverages the benefits of self-attention to operate on large point clouds and graphs with varying number of points, while guaranteeing SE(3)-equivariance for robustness. We evaluate our model on a toy $N$-body particle simulation dataset, showcasing the robustness of the predictions under rotations of the input. We further achieve competitive performance on two real-world datasets, ScanObjectNN and QM9. In all cases, our model outperforms a strong, non-equivariant attention baseline and an equivariant model without attention.
\end{abstract}

\section{Introduction}

\begin{figure}
    \centering
    \includegraphics[width=\linewidth]{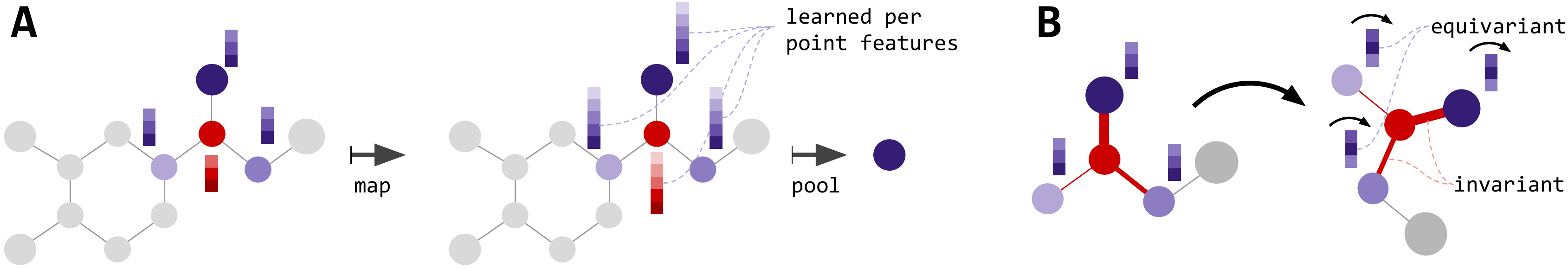}
    \caption{\textbf{A}) Each layer of the SE(3)-Transformer maps from a point cloud to a point cloud (or graph to graph) while guaranteeing equivariance. 
        For classification, this is followed by an invariant pooling layer and an MLP. \textbf{B}) In each layer, for each node, attention is performed. Here, the red node attends to its neighbours. Attention weights (indicated by line thickness) are invariant w.r.t. input rotation.
    }
    \label{fig:teaser}
\end{figure}

Self-attention mechanisms \citep{VaswaniSPUJGKP17} have enjoyed a sharp rise in popularity in recent years. Their relative implementational simplicity coupled with high efficacy on a wide range of tasks such as language modeling \citep{VaswaniSPUJGKP17}, image recognition \citep{ParmarRVBLS19}, or graph-based problems \citep{VelikovicCCRLB2017}, make them an attractive component to use. However, their generality of application means that for specific tasks, knowledge of existing underlying structure is unused. In this paper, we propose the \emph{SE(3)-Transformer} shown in \Cref{fig:teaser}, a self-attention mechanism specifically for 3D point cloud and graph data, which adheres to \emph{equivariance constraints}, improving robustness to nuisance transformations and general performance.

Point cloud data is ubiquitous across many fields, presenting itself in diverse forms such as 3D object scans \citep{uy-scanobjectnn-iccv19}, 3D molecular structures \citep{RamakrishnanDRvL14}, or $N$-body particle simulations \citep{KipfFWWZ18}. Finding neural structures which can adapt to the varying number of points in an input, while respecting the irregular sampling of point positions, is challenging. Furthermore, an important property is that these structures should be invariant to global changes in overall input pose; that is, 3D translations and rotations of the input point cloud should not affect the output. In this paper, we find that the explicit imposition of equivariance constraints on the self-attention mechanism addresses these challenges. The SE(3)-Transformer uses the self-attention mechanism as a data-dependent filter particularly suited for sparse, non-voxelised point cloud data, while respecting and leveraging the symmetries of the task at hand. 

Self-attention itself is a pseudo-linear map between sets of points. It can be seen to consist of two components: input-dependent \emph{attention weights} and an embedding of the input, called a \emph{value embedding}. In \Cref{fig:teaser}, we show an example of a molecular graph, where attached to every atom we see a value embedding vector and where the attention weights are represented as edges, with width corresponding to the attention weight magnitude. In the SE(3)-Transformer, we explicitly design the attention weights to be invariant to global pose. Furthermore, we design the value embedding to be equivariant to global pose. Equivariance generalises the translational weight-tying of convolutions. It ensures that transformations of a layer's input manifest as equivalent transformations of the output. SE(3)-equivariance in particular is the generalisation of translational weight-tying in 2D known from conventional convolutions to roto-translations in 3D.
This restricts the space of learnable functions to a subspace which adheres to the symmetries of the task and thus reduces the number of learnable parameters.
Meanwhile, it provides us with a richer form of invariance, since relative positional information between features in the input is preserved.

The works closest related to ours are tensor field networks (TFN) \citep{ThomasSKYKR18} and their voxelised equivalent, 3D steerable CNNs \cite{WeilerGWBC18}. These provide frameworks for building SE(3)-equivariant convolutional networks operating on point clouds. Employing self-attention instead of convolutions has several advantages. (1) It allows a natural handling of edge features extending TFNs to the graph setting. (2) This is one of the first examples of a nonlinear equivariant layer. In \Cref{sec:method_attention}, we show our proposed approach relieves the strong angular constraints on the filter compared to TFNs, therefore adding representational capacity. This constraint has been pointed out in the equivariance literature to limit performance severely \citep{e2cnn}. Furthermore, we provide a more efficient implementation, mainly due to a GPU accelerated version of the spherical harmonics. The TFN baselines in our experiments leverage this and use significantly scaled up architectures compared to the ones used in \cite{ThomasSKYKR18}.

Our contributions are the following:
\begin{itemize}
	\item We introduce a novel self-attention mechanism, guaranteeably invariant to global rotations and translations of its input. It is also equivariant to permutations of the input point labels.
	\item We show that the SE(3)-Transformer resolves an issue with concurrent SE(3)-equivariant neural networks, which suffer from angularly constrained filters.
	\item We introduce a \texttt{Pytorch} implementation of spherical harmonics, which is 10x faster than \texttt{Scipy} on CPU and $100 - 1000\times$ faster on GPU. This directly addresses a bottleneck of TFNs \citep{ThomasSKYKR18}. E.g., for a ScanObjectNN model, we achieve $\approx 22 \times$ speed up of the forward pass compared to a network built with SH from the \texttt{lielearn} library (see \Cref{sec:sh_accelerated}).
	\item Code available at \url{https://github.com/FabianFuchsML/se3-transformer-public}
\end{itemize}

\section{Background And Related Work}
In this section we introduce the relevant background materials on self-attention, graph neural networks, and equivariance. We are concerned with point cloud based machine learning tasks, such as object classification or segmentation. In such a task, we are given a point cloud as input, represented as a collection of $n$ coordinate vectors $\bb{x}_i \in \mbb{R}^{3}$ with optional per-point features $\bb{f}_i \in \mbb{R}^{d}$. 

\subsection{The Attention Mechanism}
The standard \emph{attention mechanism} \citep{VaswaniSPUJGKP17} can be thought of as consisting of three terms: a set of query vectors $\bb{q}_i \in \mbb{R}^p$ for $i=1,...,m$, a set of key vectors $\bb{k}_j \in \mbb{R}^{p}$ for $j=1,...,n$, and a set of value vectors $\bb{v}_j \in \mbb{R}^{r}$ for $j=1,...,n$, where $r$ and $p$ are the dimensions of the low dimensional embeddings. We commonly interpret the key $\bb{k}_j$ and the value $\bb{v}_j$ as being `attached' to the same point $j$. For a given query $\bb{q}_i$, the attention mechanism can be written as
\begin{align}
	\text{Attn}\left(\bb{q}_i, \{\bb{k}_j\}, \{\bb{v}_j\}\right) = \sum_{j=1}^n \alpha_{ij} \bb{v}_j, \qquad \alpha_{ij} = \frac{\exp(\bb{q}_i^{\top} \bb{k}_j)}{\sum_{j'=1}^n \exp(\bb{q}_i^{\top} \bb{k}_{j'})} \label{eq:self-attention}
\end{align}

where we used a softmax as a nonlinearity acting on the weights.
In general, the number of query vectors does not have to equal the number of input points \citep{SetTransformer}. In the case of \emph{self-attention} the query, key, and value vectors are embeddings of the input features, so
\begin{align}
	\bb{q} = h_Q(\bb{f}), && \bb{k} = h_K(\bb{f}), && \bb{v} = h_V(\bb{f}), \label{eq:embeddings}
\end{align}
where $\{h_Q, h_K, h_V\}$ are, in the most general case, neural networks \citep{steenkiste2018relational}. For us, query $\bb{q}_i$ is associated with a point $i$ in the input, which has a geometric location $\bb{x}_i$. Thus if we have $n$ points, we have $n$ possible queries. For query $\bb{q}_i$, we say that node $i$ \emph{attends} to all other nodes $j \neq i$.

Motivated by a successes across a wide range of tasks in deep learning such as language modeling \citep{VaswaniSPUJGKP17}, image recognition \citep{ParmarRVBLS19}, graph-based problems \citep{VelikovicCCRLB2017}, and relational reasoning \citep{steenkiste2018relational,Mohart}, 
a recent stream of work has applied forms of self-attention algorithms to point cloud data \citep{Yang2019, ShapeContextNet, SetTransformer}.
One such example is the Set Transformer \citep{SetTransformer}. When applied to object classification on ModelNet40 \citep{modelnet40}, the input to the Set Transformer are the cartesian coordinates of the points. Each layer embeds this positional information further while dynamically querying information from other points. The final per-point embeddings are downsampled and used for object classification.

\textbf{Permutation equivariance}
A key property of self-attention is \emph{permutation equivariance}. Permutations of point labels $1,...,n$ lead to permutations of the self-attention output. This guarantees the attention output does not depend arbitrarily on input point ordering. \citet{Wagstaff2019} recently showed that this mechanism can theoretically approximate \textit{all} permutation equivariant functions. The SE(3)-transformer is a special case of this attention mechanism, inheriting permutation equivariance.
However, it limits the space of learnable functions to rotation and translation equivariant ones.

\subsection{Graph Neural Networks}

Attention scales quadratically with point cloud size, so it is useful to introduce neighbourhoods: instead of each point attending to \textit{all} other points, it only attends to its nearest neighbours. Sets with neighbourhoods are naturally represented as graphs. Attention has previously been introduced on graphs under the names of intra-, self-, vertex-, or graph-attention \citep{lin2017structured, VaswaniSPUJGKP17, VelikovicCCRLB2017, hoshen2017vain, shaw2018selfattention}. These methods were unified by \citet{wang2017nonlocal} with the non-local neural network. This has the simple form
\begin{align}
    \bb{y}_i = \frac{1}{\mc{C}(\{\bb{f}_j \in \mc{N}_i\})}\sum_{j\in\mc{N}_i} w(\bb{f}_i, \bb{f}_j) h(\bb{f}_j) \label{eq:nonlocal}
\end{align}
where $w$ and $h$ are neural networks and $\mc{C}$ normalises the sum as a function of all features in the neighbourhood $\mc{N}_i$. This has a similar structure to attention, and indeed we can see it as performing attention per neighbourhood. While non-local modules do not explicitly incorporate edge-features, it is possible to add them, as done in \citet{VelikovicCCRLB2017} and \citet{hoshen2017vain}.

\subsection{Equivariance}
Given a set of transformations $T_g: \mc{V} \to \mc{V}$ for $g \in G$, where $G$ is an abstract group, a function $\phi : \mc{V} \to \mc{Y}$ is called equivariant if for every $g$ there exists a transformation $S_g: \mc{Y} \to \mc{Y}$ such that
\begin{align}
    S_g[\phi(v)] = \phi(T_g[v]) \qquad \text{for all } g\in G, v\in \mc{V} .\label{eq:equivariance-constraint}
\end{align}
The indices $g$ can be considered as parameters describing the transformation. Given a pair $(T_g, S_g)$, we can solve for the family of equivariant functions $\phi$ satisfying Equation \ref{eq:equivariance-constraint}. Furthermore, if $(T_g, S_g)$ are linear and the map $\phi$ is also linear, then a very rich and developed theory already exists for finding $\phi$~\citep{CohenW2016}. In the equivariance literature, deep networks are built from interleaved linear maps $\phi$ and equivariant nonlinearities. In the case of 3D roto-translations it has already been shown that a suitable structure for $\phi$ is a \emph{tensor field network} \citep{ThomasSKYKR18}, explained below. Note that \citet{Romero2020} recently introduced a 2D roto-translationally equivariant attention module for pixel-based image data. 

\textbf{Group Representations}
In general, the transformations $(T_g, S_g)$ are called \emph{group representations}. Formally, a group representation $\rho: G \to GL(N)$ is a map from a group $G$ to the set of $N\times N$ invertible matrices $GL(N)$. Critically $\rho$ is a \emph{group homomorphism}; that is, it satisfies the following property $\rho(g_1 g_2) = \rho(g_1)\rho(g_2)$ for all $g_1, g_2 \in G$. Specifically for 3D rotations $G=SO(3)$, we have a few interesting properties: 1) its representations are orthogonal matrices, 2) all representations can be decomposed as
\begin{align}
    \rho(g) = \bb{Q}^{\top} \left [ \bigoplus_{\ell} \bb{D}_{\ell}(g) \right ] \bb{Q}, \label{eq:so3-representations}
\end{align}
where $\bb{Q}$ is an orthogonal, $N \times N$, change-of-basis matrix \citep{chirikjian2001engineering}; each $\bb{D}_\ell$ for $\ell = 0,1,2,...$ is a $(2\ell+1) \times (2\ell + 1)$ matrix known as a Wigner-D matrix\footnote{The `D' stands for \emph{Darstellung}, German for representation}; and the $\bigoplus$ is the \emph{direct sum} or concatenation of matrices along the diagonal. The Wigner-D matrices are \emph{irreducible representations} of SO(3)---think of them as the `smallest' representations possible. 
Vectors transforming according to $\bb{D}_\ell$ (i.e.\ we set $\bb{Q} = \bb{I}$), are called \emph{type}-$\ell$ vectors. Type-0 vectors are invariant under rotations and type-1 vectors rotate according to 3D rotation matrices. Note, type-$\ell$ vectors have length $2\ell+1$. They can be stacked, forming a feature vector $\bb{f}$ transforming according to \Cref{eq:so3-representations}. 

\textbf{Tensor Field Networks}
\label{sec:background-tfn}
Tensor field networks (TFN) \citep{ThomasSKYKR18} are neural networks, which map point clouds to point clouds under the constraint of SE(3)-equivariance, the group of 3D rotations and translations. For point clouds, the input is a vector field $\bb{f}: \mbb{R}^3 \to \mbb{R}^d$ of the form
\begin{align}
    \bb{f}(\bb{x}) = \sum_{j=1}^N \bb{f}_j \delta(\bb{x} - \bb{x}_j), \label{eq:point-cloud}
\end{align}
where $\delta$ is the Dirac delta function, $\{\bb{x}_j\}$ are the 3D point coordinates and $\{\bb{f}_j\}$ are point features, representing such quantities as atomic number or point identity. For equivariance to be satisfied, the features of a TFN transform under \Cref{eq:so3-representations}, where $\bb{Q}=\bb{I}$. Each $\bb{f}_j$ is a concatenation of vectors of different \emph{types}, where a subvector of type-$\ell$ is written $\bb{f}_j^\ell$. A TFN layer computes the convolution of a continuous-in-space, learnable weight kernel $\bb{W}^{\ell k}: \mathbb{R}^3 \to \mathbb{R}^{(2\ell+1)\times(2k+1)}$ from type-$k$ features to type-$\ell$ features. The type-$\ell$ output of the TFN layer at position $\bb{x}_i$ is
\begin{align}
    \bb{f}_{\text{out},i}^{\ell} = \sum_{k\geq 0} \underbrace{\int \bb{W}^{\ell k}(\bb{x}'-\bb{x}_i) \bb{f}_{\text{in}}^k(\bb{x}') \, \mathrm{d}\bb{x}'}_{k \to \ell \text{ convolution}} = \sum_{k\geq 0}\sum_{j=1}^n \underbrace{\bb{W}^{\ell k}(\bb{x}_j - \bb{x}_i) \bb{f}_{\text{in},j}^k,}_{\text{node $j \to$ node $i$ message}} \label{eq:tf-layer}
\end{align}
We can also include a sum over input channels, but we omit it here. \citet{WeilerGWBC18, ThomasSKYKR18} and \citet{kondor2018nbody} showed that the kernel $\bb{W}^{\ell k}$ lies in the span of an equivariant basis $\{\bb{W}_J^{\ell k}\}_{J=|k-\ell|}^{k+\ell}$. The kernel is a linear combination of these basis kernels, where the $J$\textsuperscript{th} coefficient is a learnable function $\varphi_J^{\ell k}: \mathbb{R}_{\geq 0} \to \mathbb{R}$ of the radius $\|\bb{x}\|$. Mathematically this is
\begin{align}
    \bb{W}^{\ell k}(\bb{x}) = \sum_{J = |k-\ell|}^{k+\ell} \varphi_J^{\ell k}(\|\bb{x}\|) \bb{W}_{J}^{\ell k}(\bb{x}), \qquad \text{where } \bb{W}_{J}^{\ell k}(\bb{x}) = \sum_{m=-J}^J Y_{Jm}(\bb{x} / \|\bb{x}\|) \bb{Q}_{Jm}^{\ell k}. \label{eq:equivariant-kernel}
\end{align}
Each basis kernel $\bb{W}_{J}^{\ell k}: \mathbb{R}^3 \to \mathbb{R}^{(2\ell+1)\times(2k+1)}$ is formed by taking a linear combination of Clebsch-Gordan matrices $\bb{Q}_{Jm}^{\ell k}$ of shape $(2\ell+1)\times(2k+1)$, where the $J,m$\textsuperscript{th} linear combination coefficient is the $m$\textsuperscript{th} dimension of the $J$\textsuperscript{th} spherical harmonic $Y_J: \mathbb{R}^3 \to \mathbb{R}^{2J+1}$. Each basis kernel $\bb{W}_{J}^{\ell k}$ completely constrains the form of the learned kernel in the angular direction, leaving the only learnable degree of freedom in the radial direction. Note that $\bb{W}_{J}^{\ell k}(\bb{0}) \neq \bb{0}$ only when $k=\ell$ and $J=0$, which reduces the kernel to a scalar $w$ multiplied by the identity, $\bb{W}^{\ell\ell} = w^{\ell\ell}\bb{I}$, referred to as \emph{self-interaction}~\citep{ThomasSKYKR18}. As such we can rewrite the TFN layer as
\begin{align}
    \bb{f}_{\text{out},i}^{\ell} = \underbrace{w^{\ell\ell} \bb{f}_{\text{in},i}^{\ell}}_{\text{self-interaction}} + \sum_{k\geq 0}\sum_{j\neq i}^n \bb{W}^{\ell k}(\bb{x}_j - \bb{x}_i) \bb{f}_{\text{in},j}^k, \label{eq:tf-layer2}
\end{align}
\Cref{eq:tf-layer} and \Cref{eq:tf-layer2} present the convolution in message-passing form, where messages are aggregated from all nodes and feature types. They are also a form of nonlocal graph operation as in \Cref{eq:nonlocal}, where the weights are functions on edges and the features $\{\bb{f}_i\}$ are node features. We will later see how our proposed attention layer unifies aspects of convolutions and graph neural networks. 

\begin{figure}
    \centering
    \includegraphics[width=\linewidth]{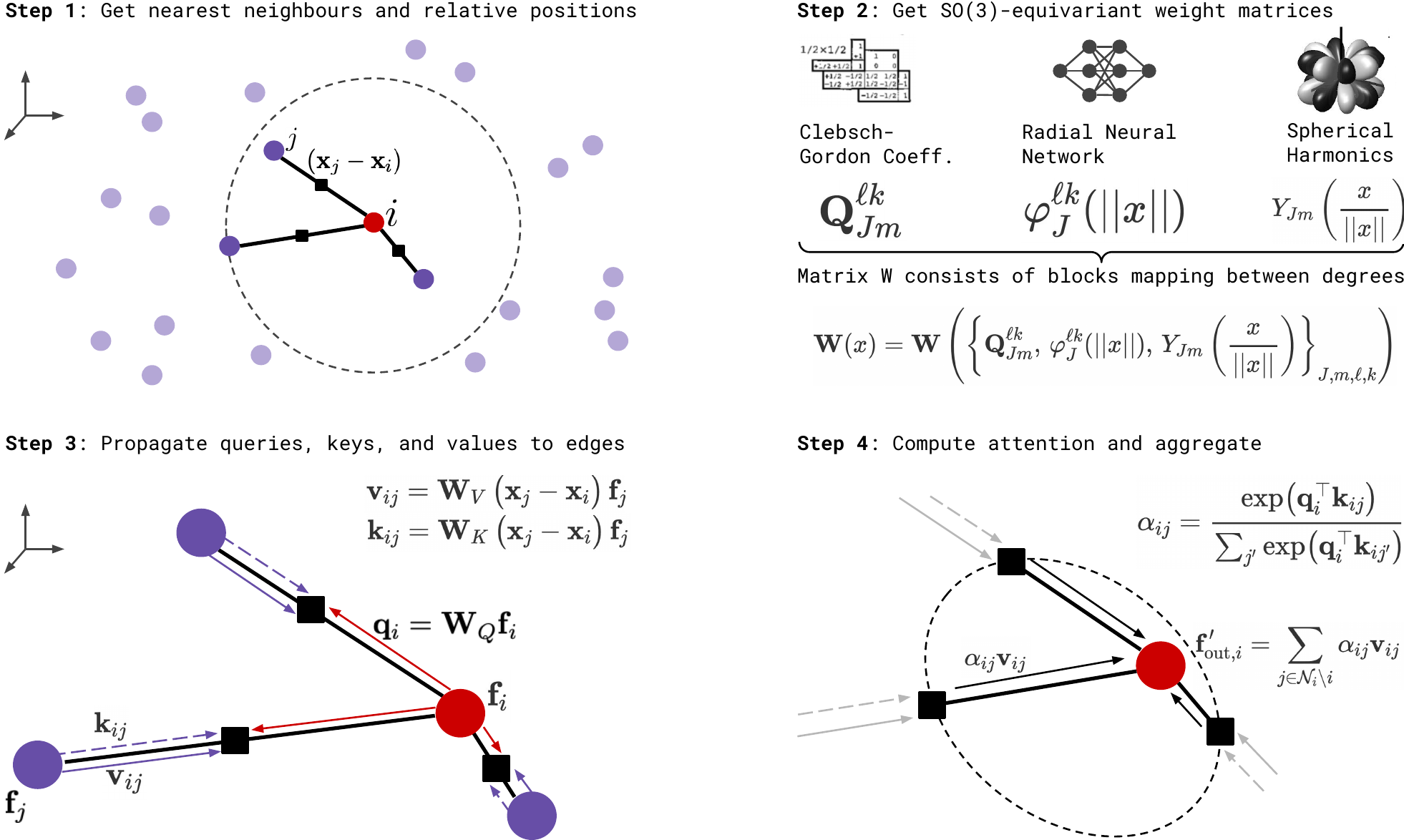}
    \caption{Updating the node features using our equivariant attention mechanism in four steps. A more detailed description, especially of step 2, is provided in the Appendix. Steps 3 and 4 visualise a graph network perspective: features are passed from nodes to edges to compute keys, queries and values, which depend both on features and relative positions in a rotation-equivariant manner.}
    \label{fig:4steps}
    \vspace{-4mm}
\end{figure}

\section{Method}
Here, we present the \emph{SE(3)-Transformer}. The layer can be broken down into a procedure of steps as shown in \Cref{fig:4steps}, which we describe in the following section. These are the construction of a graph from a point cloud, the construction of equivariant edge functions on the graph, how to propagate SE(3)-equivariant messages on the graph, and how to aggregate them. We also introduce an alternative for the self-interaction layer, which we call \emph{attentive self-interaction}.

\subsection{Neighbourhoods}
Given a point cloud $\{(\bb{x}_i, \bb{f}_i)\}$, we first introduce a collection of neighbourhoods $\mc{N}_i \subseteq \{1,...,N\}$, one centered on each point $i$. These neighbourhoods are computed either via the nearest-neighbours methods or may already be defined. For instance, molecular structures have neighbourhoods defined by their bonding structure. Neighbourhoods reduce the computational complexity of the attention mechanism from quadratic in the number of points to linear. The introduction of neighbourhoods converts our point cloud into a graph. This step is shown as Step 1 of \Cref{fig:4steps}.

\subsection{The SE(3)-Transformer}
\label{sec:method_attention}
The SE(3)-Transformer itself consists of three components. These are 1) edge-wise attention weights $\alpha_{ij}$, constructed to be SE(3)-invariant on each edge $ij$, 2) edge-wise SE(3)-equivariant value messages, propagating information between nodes, as found in the TFN convolution of \Cref{eq:tf-layer}, and 3) a linear/attentive self-interaction layer. Attention is performed on a per-neighbourhood basis as follows:
\begin{align}
\label{eq:att_equi}
    \bb{f}_{\text{out},i}^\ell = \underbrace{\bb{W}_V^{\ell \ell} \bb{f}_{\text{in},i}^{\ell}}_{\text{ \textcircled{\raisebox{-0.9pt}{3}} self-interaction}} + \sum_{k \geq 0} \sum_{j \in \mc{N}_i \setminus i} \underbrace{\alpha_{ij}}_{\text{\textcircled{\raisebox{-0.6pt}{1}} attention}} \underbrace{\bb{W}_V^{\ell k}(\bb{x}_j - \bb{x}_i) \bb{f}_{\text{in},j}^k}_{\text{\textcircled{\raisebox{-0.9pt}{2}} value message}}.
\end{align}
These components are visualised in \Cref{fig:4steps}. If we remove the attention weights then we have a tensor field convolution,
and if we instead remove the dependence of $\bb{W}_V$ on $(\bb{x}_j - \bb{x}_i)$, we have a conventional attention mechanism.
Provided that the attention weights $\alpha_{ij}$ are invariant, \Cref{eq:att_equi} is equivariant to SE(3)-transformations. This is because it is just a linear combination of equivariant value messages. Invariant attention weights can be achieved with a dot-product attention structure shown in \Cref{eq:att_equi_weights}. This mechanism consists of a normalised inner product between a query vector $\bb{q}_i$ at node $i$ and a set of key vectors $\{\bb{k}_{ij}\}_{j \in \mc{N}_i}$ along each edge $ij$ in the neighbourhood $\mc{N}_i$ where
\begin{align}
\label{eq:att_equi_weights}
    \alpha_{ij} = \frac{\exp ( \bb{q}_i^{\top} \bb{k}_{ij} ) }{\sum_{j' \in \mc{N}_i\setminus i} \exp ( \bb{q}_i^{\top} \bb{k}_{ij'})}, \;\;\; \bb{q}_i = \bigoplus_{\ell \geq 0} \sum_{k \geq 0}\bb{W}_Q^{\ell k}\bb{f}_{\text{in},i}^k, \;\;\; \bb{k}_{ij} = \bigoplus_{\ell \geq 0} \sum_{k\geq 0} \bb{W}_K^{\ell k}(\bb{x}_j - \bb{x}_i) \bb{f}_{\text{in},j}^k.
\end{align}
$\bigoplus$ is the direct sum, i.e.\ vector concatenation in this instance. The linear embedding matrices $\bb{W}_Q^{\ell k}$ and $\bb{W}_K^{\ell k}(\bb{x}_j - \bb{x}_i)$ are of TFN type (c.f.\ \Cref{eq:equivariant-kernel}). The attention weights $\alpha_{ij}$ are invariant for the following reason. If the input features $\{ \bb{f}_{\text{in},j} \}$ are SO(3)-equivariant, then the query $\bb{q}_i$ and key vectors $\{\bb{k}_{ij}\}$ are also SE(3)-equivariant, since the linear embedding matrices are of TFN type. The inner product of SO(3)-equivariant vectors, transforming under the same representation $\bb{S}_g$ is invariant, since if $\bb{q} \mapsto \bb{S}_g \bb{q}$ and $\bb{k} \mapsto \bb{S}_g \bb{k}$, then $\bb{q}^\top \bb{S}_g^\top \bb{S}_g \bb{k} = \bb{q}^\top \bb{k}$, because of the orthonormality of representations of SO(3), mentioned in the background section. We follow the common practice from the self-attention literature \citep{VaswaniSPUJGKP17,SetTransformer}, and chosen a softmax nonlinearity to normalise the attention weights to unity, but in general any nonlinear function could be used.

\textbf{Aside: Angular Modulation}
The attention weights add extra degrees of freedom to the TFN kernel in the angular direction. This is seen when \Cref{eq:att_equi} is viewed as a convolution with a data-dependent kernel $\alpha_{ij} \bb{W}_V^{\ell k}(\bb{x})$. In the literature, SO(3) equivariant kernels are decomposed as a sum of products of learnable radial functions $\varphi_J^{\ell k}(\|\bb{x}\|)$ and non-learnable angular kernels $\bb{W}_J^{\ell k}(\bb{x} / \|\bb{x}\|)$ (c.f.\ \Cref{eq:equivariant-kernel}). The fixed angular dependence of $\bb{W}_J^{\ell k}(\bb{x} / \|\bb{x}\|)$ is a strange artifact of the equivariance condition in noncommutative algebras and while necessary to guarantee equivariance, it is seen as overconstraining the expressiveness of the kernels. Interestingly, the attention weights $\alpha_{ij}$ introduce a means to modulate the angular profile of $\bb{W}_J^{\ell k}(\bb{x} / \|\bb{x}\|)$, while maintaining equivariance.

\textbf{Channels, Self-interaction Layers, and Non-Linearities}
Analogous to conventional neural networks, the SE(3)-Transformer can straightforwardly be extended to multiple channels per representation degree $\ell$, so far omitted for brevity. This sets the stage for self-interaction layers. The attention layer (c.f.\ \Cref{fig:4steps} and circles 1 and 2 of \Cref{eq:att_equi}) aggregates information over nodes and input representation degrees $k$. In contrast, the self-interaction layer (c.f.\ circle 3 of \Cref{eq:att_equi}) exchanges information solely between features of the same degree and within one node---much akin to 1x1 convolutions in CNNs. Self-interaction is an elegant form of learnable skip connection, transporting information from query point $i$ in layer $L$ to query point $i$ in layer $L+1$. This is crucial since, in the SE(3)-Transformer, points do not attend to themselves. In our experiments, we use two different types of self-interaction layer: (1) linear and (2) attentive, both of the form
\begin{align}
    \bb{f}_{\text{out},i,c'}^\ell &= \sum_{c} w_{i,c'c}^{\ell\ell} \bb{f}_{\text{in},i,c}^\ell \label{eq:self-interaction}.
\end{align}
\textbf{Linear:} Following \citet{schnet}, output channels are a learned linear combination of input channels using one set of weights $w_{i,c'c}^{\ell\ell} = w_{c'c}^{\ell\ell}$ per representation degree, shared across all points. As proposed in \citet{ThomasSKYKR18}, this is followed by a norm-based non-linearity. 

\textbf{Attentive}: We propose an extension of linear self-interaction, \emph{attentive self-interaction}, combining self-interaction and nonlinearity. We replace the learned scalar weights $w_{c'c}^{\ell\ell}$ with attention weights output from an MLP, shown in \Cref{eq:attentive-self-interaction} ($\bigoplus$ means concatenation.). These weights are SE(3)-invariant due to the invariance of inner products of features, transforming under the same representation.
\begin{align}
w_{i,c'c}^{\ell\ell} = \text{MLP}\left( \bigoplus_{c,c'}\bb{f}_{\text{in},i,c'}^{\ell\top}\bb{f}_{\text{in},i,c}^\ell \right)
\label{eq:attentive-self-interaction}
\end{align}

\subsection{Node and Edge Features}
Point cloud data often has information attached to points (node-features) and connections between points (edge-features), which we would both like to pass as inputs into the first layer of the network.
Node information can directly be incorporated via the tensors $\bb{f}_j$ in \Cref{eq:point-cloud,,eq:att_equi}.
For incorporating edge information, note that $\bb{f}_j$ is part of multiple neighbourhoods. One can replace $\bb{f}_j$ with $\bb{f}_{ij}$ in \Cref{eq:att_equi}. Now, $\bb{f}_{ij}$ can carry different information depending on which neighbourhood $\mc{N}_i$ we are currently performing attention over. In other words, $\bb{f}_{ij}$ can carry information both about node $j$ but also about edge $ij$. Alternatively, if the edge information is scalar, it can be incorporated into the weight matrices $\bb{W}_V$ and $\bb{W}_K$ as an input to the radial network (see step 2 in \Cref{fig:4steps}).

\input{experiments}

\input{conclusion}

\input{impact}

\section*{Acknowledgements and Funding Disclosure}
We would like to express our gratitude to the Bosch Center for Artificial Intelligence and Konincklijke Philips N.V. for funding our work and contributing to open research by publishing our paper. Fabian Fuchs worked on this project while on a research sabbatical at the Bosch Center for Artificial Intelligence. His PhD is funded by Kellogg College Oxford and the EPSRC AIMS Centre
for Doctoral Training at Oxford University.

\newpage
\bibliographystyle{plainnat}
\bibliography{example_paper}

\input{appendix}

\end{document}

%% file: experiments.tex
\begin{figure}[b]
\vspace{-2mm}
    \centering
    \begin{subfigure}[c]{0.47\linewidth}
        \centering
        \includegraphics[width=\linewidth]{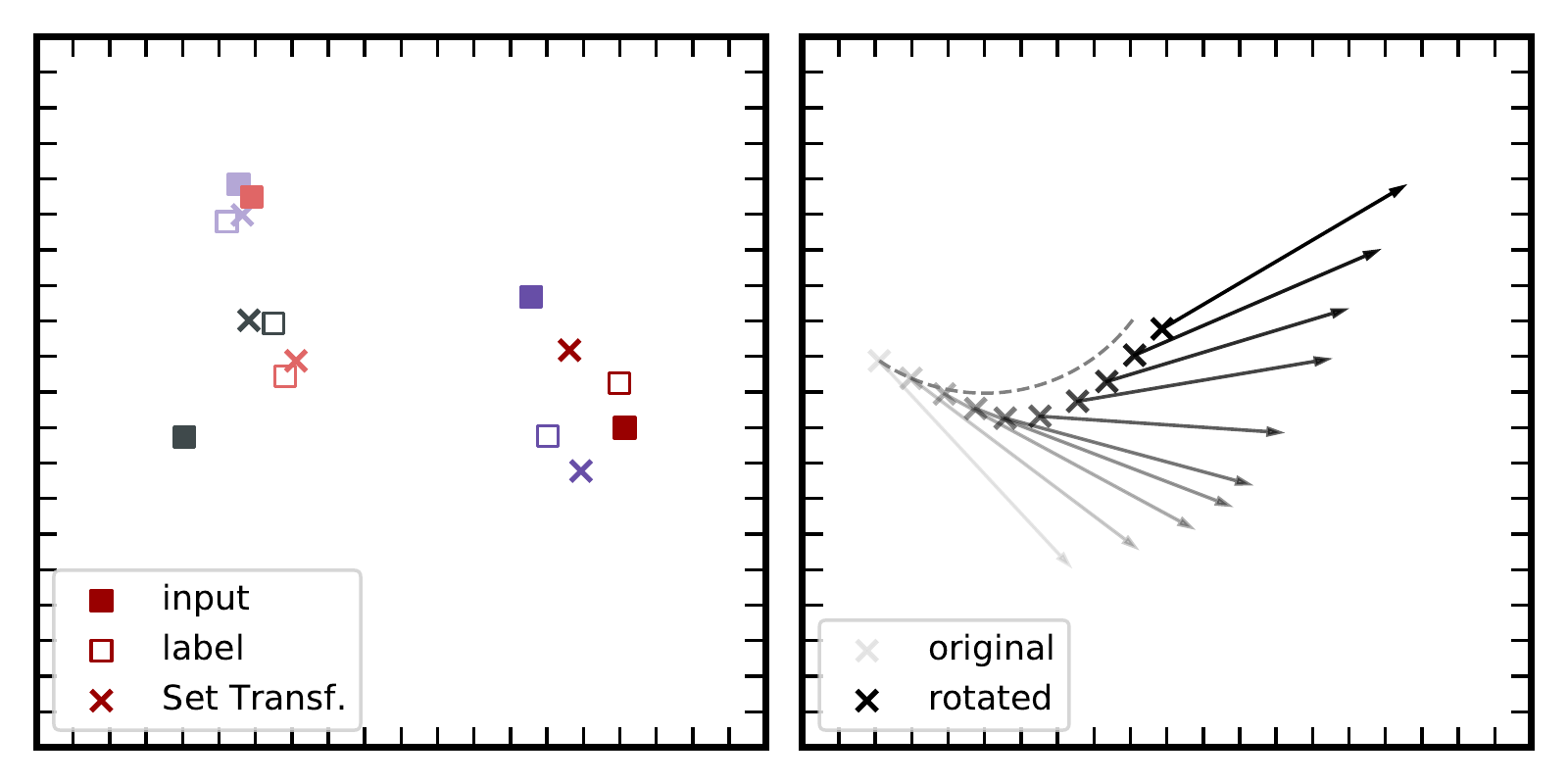}
        \vspace{-6mm}
        \subcaption{Set Transformer}
        \label{fig:nri_rotation_example_A}
    \end{subfigure}
    \hspace*{\fill}
    \begin{subfigure}[c]{0.47\linewidth}
        \centering
        \includegraphics[width=\linewidth]{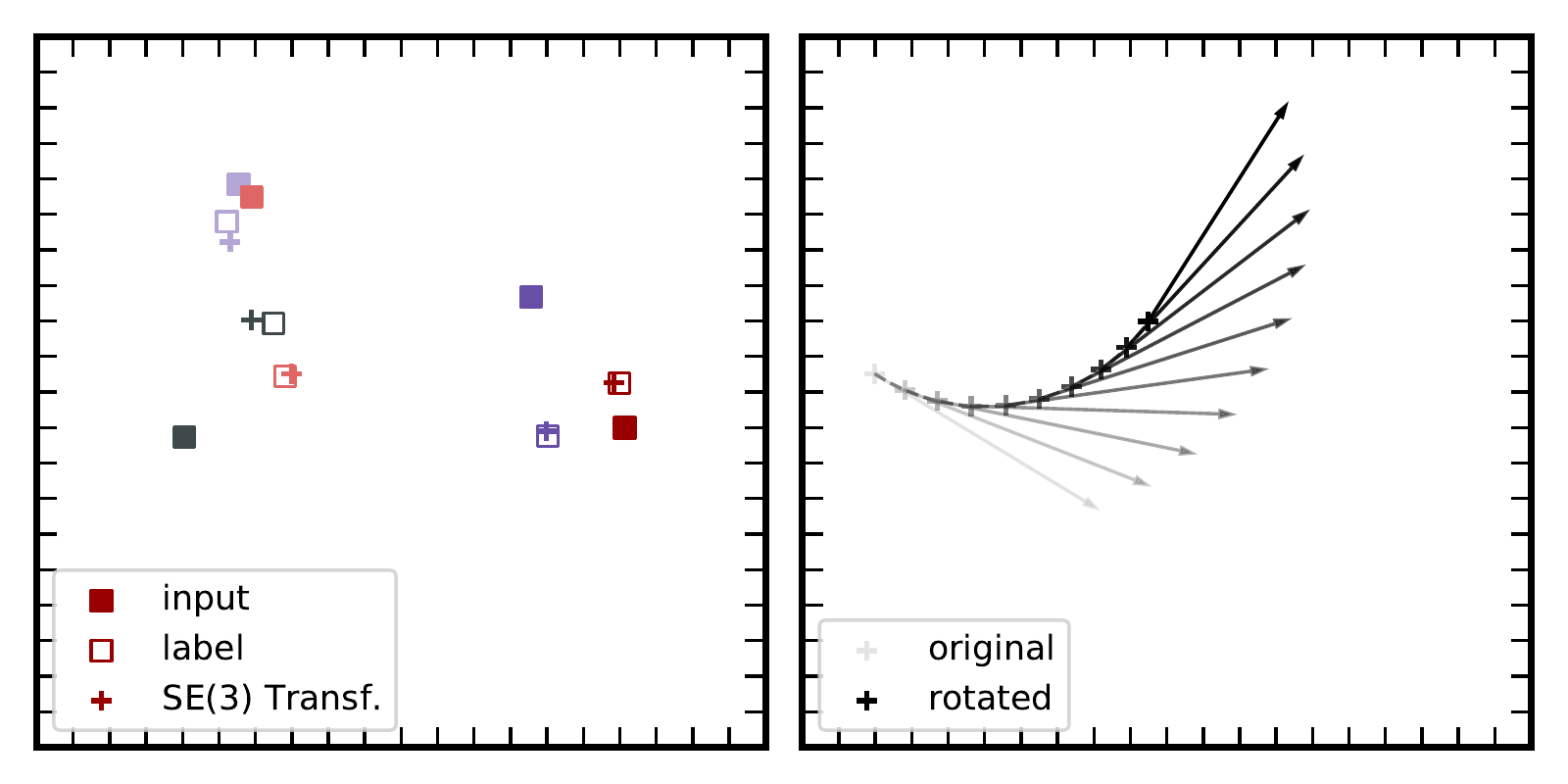}
        \vspace{-6mm}
        \subcaption{SE(3)-Transformer}
        \label{fig:nri_rotation_example_B}
    \end{subfigure}
    \vspace{-1mm}
    \caption{A model based on conventional self-attention (left) and our rotation-equivariant version (right) predict future locations and velocities in a 5-body problem. The respective left-hand plots show input locations at time step $t=0$, ground truth locations at $t=500$, and the respective predictions. The right-hand plots show predicted locations and velocities for rotations of the input in steps of 10 degrees. The dashed curves show the predicted locations of a perfectly equivariant model.
    }
    \label{fig:nri_rotation_example}
\end{figure}
\section{Experiments}
We test the efficacy of the SE(3)-Transformer on three datasets, each testing different aspects of the model. The N-body problem is an equivariant task: rotation of the input should result in rotated predictions of locations and velocities of the particles. Next, we evaluate on a real-world object classification task. Here, the network is confronted with large point clouds of noisy data with symmetry only around the gravitational axis. Finally, we test the SE(3)-Transformer on a molecular property regression task, which shines light on its ability to incorporate rich graph structures. We compare to publicly available, state-of-the-art results as well as a set of our own baselines. Specifically, we compare to the Set-Transformer \citep{SetTransformer}, a non-equivariant attention model, and Tensor Field Networks \citep{ThomasSKYKR18}, which is similar to SE(3)-Transformer but does not leverage attention. 

Similar to \cite{Sosnovik20, Worrall19}, we measure the exactness of equivariance by applying uniformly sampled SO(3)-transformations to input and output. The distance between the two, averaged over samples, yields the equivariance error. Note that, unlike in \citet{Sosnovik20}, the error is not squared:
\begin{align}
    \Delta_{EQ}=\left\| L_{s} \Phi(f)-\Phi L_{s}(f)\right\|_{2} /\left\| L_{s} \Phi(f) \|_{2}\right.
\end{align}

\subsection{N-Body Simulations}

In this experiment, we use an adaptation of the dataset from \citet{KipfFWWZ18}. Five particles each carry either a positive or a negative charge and exert repulsive or attractive forces on each other. The input to the network is the position of a particle in a specific time step, its velocity, and its charge. The task of the algorithm is then to predict the relative location and velocity 500 time steps into the future. We deliberately formulated this as a regression problem to avoid the need to predict multiple time steps iteratively. Even though it certainly is an interesting direction for future research to combine equivariant attention with, e.g., an LSTM, our goal here was to test our core contribution and compare it to related models. This task sets itself apart from the other two experiments by not being invariant but equivariant: When the input is rotated or translated, the output changes respectively (see \Cref{fig:nri_rotation_example}).

We trained an SE(3)-Transformer with 4 equivariant layers, each followed by an attentive self-interaction layer (details are provided in the Appendix). \Cref{tab:RI} shows quantitative results. 
Our model outperforms both an attention-based, but not rotation-equivariant approach (Set Transformer) and a equivariant approach which does not levarage attention (Tensor Field). The equivariance error shows that our approach is indeed fully rotation equivariant up to the precision of the computations.

\begin{table}[]
{\caption{Predicting future locations and velocities in an electron-proton simulation.}
\label{tab:RI}}
\resizebox{\columnwidth}{!}{
\begin{tabular}{llrrrrr}
    \toprule
       &                     & Linear               & DeepSet \cite{Zaheer2017}              & Tensor Field \cite{ThomasSKYKR18}         & Set Transformer \cite{SetTransformer}      & \textbf{SE(3)-Transformer} \\
       \midrule
      & MSE $x$             & $0.0691$             & $0.0639$             & $0.0151$             & $0.0139$             & $\bm{0.0076}$              \\
       \textbf{Position} & std                 & -                    & $0.0086$             & $0.0011$             & $0.0004$             & $0.0002$              \\
       & $\Delta_{EQ}$       & -                    & $0.038$              & $1.9 \cdot 10^{-7}$  & $0.167$              & $3.2 \cdot 10^{-7}$   \\
       \midrule
      & MSE $v$             & $0.261$             & $0.246$              & $0.125$              & $0.101$              & $\bm{0.075}$              \\
       \textbf{Velocity} & std                 & -                    & $0.017$              & $0.002$              & $0.004$              & $0.001$               \\
       & $\Delta_{EQ}$       & -                    & $1.11$               & $5.0 \cdot 10^{-7}$  & $0.37$               & $6.3 \cdot 10^{-7}$   \\ 
       \bottomrule
\end{tabular}}
\vspace{-3mm}
\end{table}

\subsection{Real-World Object Classification on ScanObjectNN}

\begin{figure}[t]
    \centering
    \begin{subfigure}[c]{0.45\linewidth}
        \centering
        \includegraphics[width=\linewidth]{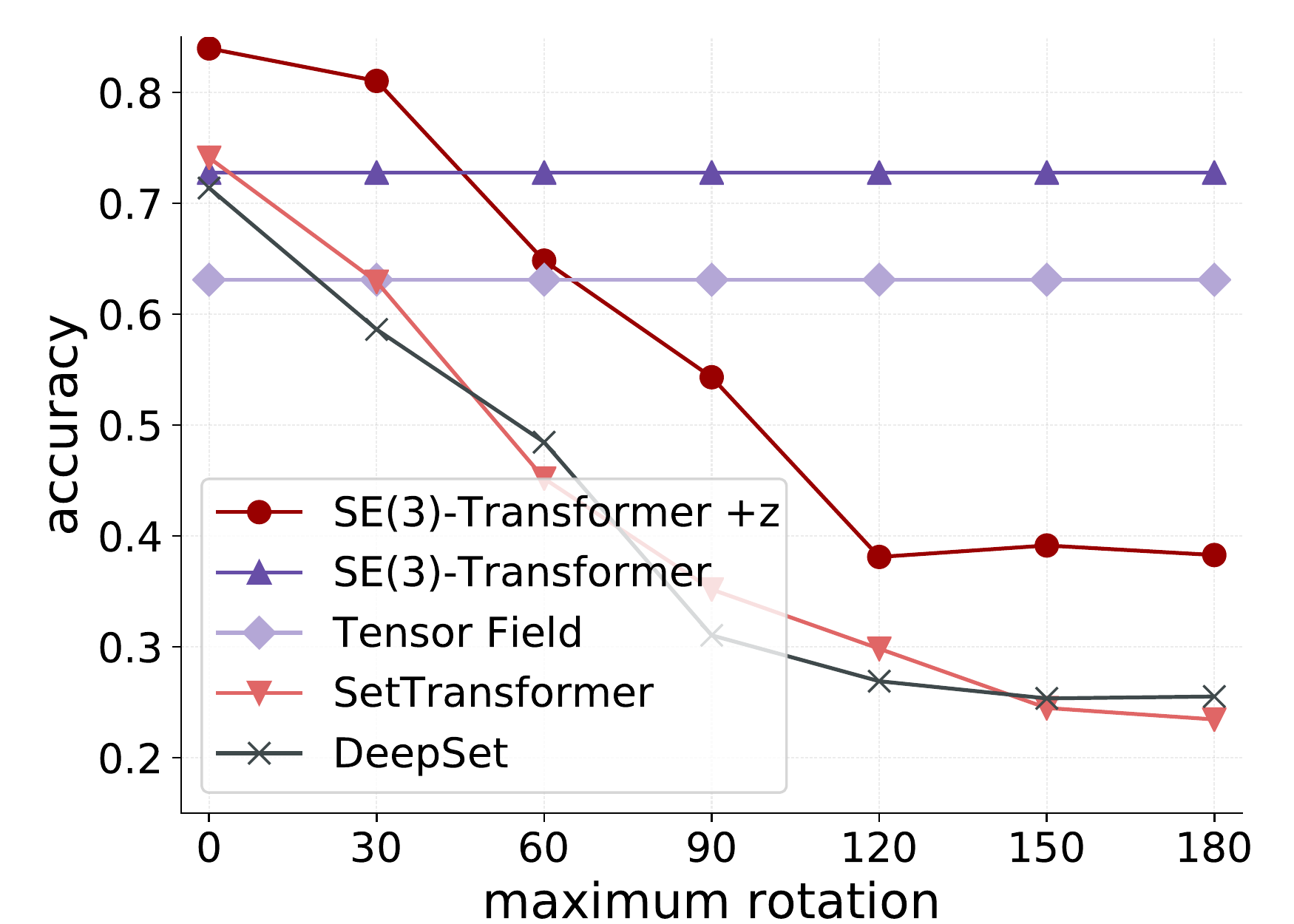}
        \subcaption{Training \textit{without} data augmentation.}
        \label{fig:scanobject_rotate_eval}
    \end{subfigure}
    \hspace*{\fill}
    \begin{subfigure}[c]{0.45\linewidth}
        \centering
        \includegraphics[width=\linewidth]{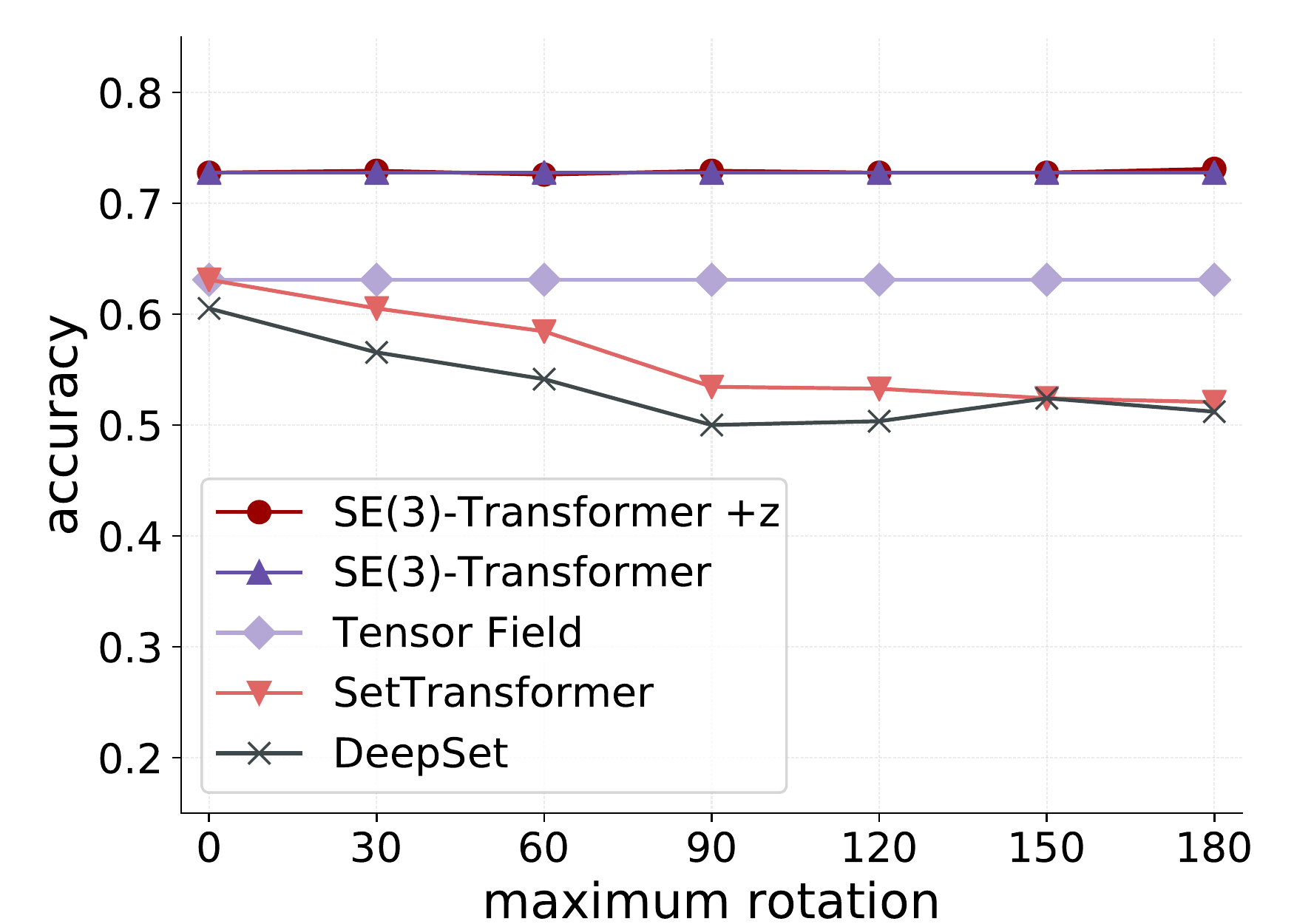}
        \subcaption{Training \textit{with} data augmentation.}
        \label{fig:scanobject_rotate_both}
    \end{subfigure}
    \caption{
        ScanObjectNN: $x$-axis shows data augmentation on the test set. The $x$-value corresponds to the maximum rotation around a random axis in the $x$-$y$-plane.
        If both training and test set are not rotated ($x=0$ in \textbf{a}), breaking the symmetry of the SE(3)-Transformer by providing the $z$-component of the coordinates as an additional, scalar input improves the performance significantly. Interestingly, the model learns to ignore the additional, symmetry-breaking input when the training set presents a rotation-invariant problem (strongly overlapping dark red circles and dark purple triangles in \textbf{b}).
    }
    \label{fig:scanobject_rotate}
\end{figure}

Object classification from point clouds is a well-studied task. Interestingly, the vast majority of current neural network methods work on scalar coordinates without incorporating vector specific inductive biases. Some recent works explore rotation invariant point cloud classification \cite{You2019,zhang2019rotation,ClusterNet,RaoFractal}. Our method sets itself apart by using roto-translation equivariant layers acting directly on the point cloud without prior projection onto a sphere \cite{RaoFractal,You2019,cohen2018spherical}. This allows for weight-tying and increased sample efficiency while transporting information about local and global orientation through the network - analogous to the translation equivariance of CNNs on 2D images. To test our method, we choose ScanObjectNN, a recently introduced dataset for real-world object classification. 
The benchmark provides point clouds of 2902 objects across 15 different categories. We only use the coordinates of the points as input and object categories as training labels. We train an SE(3)-Transformer with 4 equivariant layers with linear self-interaction followed by max-pooling and an MLP. Interestingly, the task is not fully rotation invariant, in a statistical sense, as the objects are aligned with respect to the gravity axis. This results in a performance loss when deploying a fully SO(3) invariant model (see \Cref{fig:scanobject_rotate_eval}). In other words: when looking at a new object, it helps to know where `up' is.
We create an SO(2) invariant version of our algorithm by additionally feeding the $z$-component as an type-0 field and the $x$, $y$ position as an additional type-1 field (see Appendix). We dub this model \textit{SE(3)-Transformer +z}. This way, the model can `learn' which symmetries to adhere to by suppressing and promoting different inputs (compare \Cref{fig:scanobject_rotate_eval} and \Cref{fig:scanobject_rotate_both}). In \Cref{tab:Scan}, we compare our model to the current state-of-the-art in object classification\footnote{At time of submission, PointGLR was a recently published preprint \citep{PointGLR}. The performance of the following models was taken from the official benchmark of the dataset as of June 4th, 2020 (\url{https://hkust-vgd.github.io/benchmark/}): 3DmFV \citep{3dmfv}, PointNet \citep{Pointnet}, SpiderCNN \citep{SpiderCNN}, PointNet++ \citep{Pointnetpp}, DGCN \citep{DGCNN}.}. Despite the dataset not playing to the strengths of our model (full SE(3)-invariance) and a much lower number of input points, the performance is competitive with models specifically designed for object classification - PointGLR \citep{PointGLR}, for instance, is pre-trained on the larger ModelNet40 dataset \citep{modelnet40}. For a discussion of performance vs. number of input points used, see \Cref{sec:app_n_points}.

\begin{table}[t]
\vspace{-1mm}
{\caption{Classification accuracy on the 'object only' category of the ScanObjectNN dataset\textsuperscript{4}. The performance of the SE(3)-Transformer is averaged over 5 runs (standard deviation 0.7\%).}
\label{tab:Scan}}
\centering
\resizebox{0.98\columnwidth}{!}{
\vspace{2mm}
\begin{tabular}{lrrrrrrrrrrrrr}
\toprule
 & \mcrot{1}{l}{30}{Tensor Field} & \mcrot{1}{l}{30}{DeepSet} & \mcrot{1}{l}{30}{\textbf{SE(3)-Transf.}} & \mcrot{1}{l}{30}{3DmFV} & \mcrot{1}{l}{30}{Set Transformer} & \mcrot{1}{l}{30}{PointNet} & \mcrot{1}{l}{30}{SpiderCNN} & \mcrot{1}{l}{30}{Tensor Field +z} & \mcrot{1}{l}{30}{PointNet++} & \mcrot{1}{l}{30}{\textbf{SE(3)-Transf.+z}} & \mcrot{1}{l}{30}{PointCNN} & \mcrot{1}{l}{30}{DGCNN} & \mcrot{1}{l}{30}{PointGLR}\\
\midrule
\vspace{1mm}
No. Points & 128 & 1024 & \textbf{128} & 1024 & 1024 & 1024 & 1024 & 128 & 1024 & \textbf{128}  & 1024  & 1024  & 1024 \\
\vspace{1mm}
Accuracy & 63.1\% & 71.4\% & \textbf{72.8} \% & 73.8\% & 74.1\% & 79.2\% & 79.5\% & 81.0\% & 84.3\% & \textbf{85.0\%} & 85.5\% & 86.2\% & 87.2\% \\
\bottomrule
\end{tabular}
}
\vspace{-2mm}
\end{table}

\newpage
\subsection{QM9}

\begin{wraptable}[10]{R}{0.65\linewidth} 
\vspace{-8mm}
{\caption{QM9 Mean Absolute Error. Top: Non-equivariant models. Bottom: Equivariant models. SE(3)-Tr. is averaged over 5 runs.}
\label{tab:QM9}}
\resizebox{0.65\columnwidth}{!}{
\begin{tabular}{lrrrrrr}
    \toprule
    \textsc{Task}   & $\alpha$ & $\Delta \varepsilon$ & $\varepsilon_\text{HOMO}$ & $\varepsilon_\text{LUMO}$ & $\mu$ & $C_\nu$ \\
    \textsc{Units}  & $\text{bohr}^3$ & meV & meV & meV & D & cal/mol K  \\
    \midrule 
    WaveScatt~\citep{HirnMP17}   & .160  & 118   & 85    & 76    & .340  & .049  \\
    NMP~\citep{Gilmer2017}        & .092  & 69    & 43    & 38    & .030  & .040   \\
    SchNet~\citep{schnet}     & .235  & 63    & 41    & 34    & .033  & .033   \\
    \midrule
    Cormorant~\cite{anderson2019cormorant}   & .085  & 61    & 34    & 38    & .038  & .026 \\
    LieConv(T3)~\citep{Finzi2020} & .084 & 49   & 30   & 25    & .032  & .038 \\
    TFN~\citep{ThomasSKYKR18} & .223  & 58    & 40    & 38    & .064  & .101 \\
    \textbf{SE(3)-Transformer}          & .142$\pm$.002  & 53.0$\pm$0.3    & 35.0$\pm$.9    & 33.0$\pm$.7    & .051$\pm$.001  & .054$\pm$.002 \\
    \bottomrule
\end{tabular}
}
\end{wraptable}

The QM9 regression dataset \citep{RamakrishnanDRvL14} is a publicly available chemical property prediction task. There are 134k molecules with up to 29 atoms per molecule. Atoms are represented as a 5 dimensional one-hot node embeddings in a molecular graph connected by 4 different chemical bond types  (more details in Appendix). `Positions' of each atom are provided. We show results on the test set of \citet{anderson2019cormorant} for 6 regression tasks in \Cref{tab:QM9}. Lower is better. The table is split into non-equivariant (top) and equivariant models (bottom). Our nearest models are Cormorant and TFN (own implementation). We see that while not state-of-the-art, we offer competitive performance, especially against Cormorant and TFN, which transform under irreducible representations of SE(3) (like us), unlike LieConv(T3), using a left-regular representation of SE(3), which may explain its success.

%% file: conclusion.tex
\section{Conclusion}
\vspace{-1mm}
We have presented an attention-based neural architecture designed specifically for point cloud data. This architecture is guaranteed to be robust to rotations and translations of the input, obviating the need for training time data augmentation and ensuring stability to arbitrary choices of coordinate frame. The use of self-attention allows for anisotropic, data-adaptive filters, while the use of neighbourhoods enables scalability to large point clouds. We have also introduced the interpretation of the attention mechanism as a data-dependent nonlinearity, adding to the list of equivariant nonlinearties which we can use in equivariant networks. Furthermore, we provide code for a speed up of spherical harmonics computation of up to 3 orders of magnitudes. This speed-up allowed us to train significantly larger versions of both the SE(3)-Transformer and the Tensor Field network \citep{ThomasSKYKR18} and to apply these models to real-world datasets.

Our experiments showed that adding attention to a roto-translation-equivariant model consistently led to higher accuracy and increased training stability. Specifically for large neighbourhoods, attention proved essential for model convergence. On the other hand, compared to convential attention, adding the equivariance constraints also increases performance in all of our experiments while at the same time providing a mathematical guarantee for robustness with respect to rotations of the input data.



%% file: impact.tex
\section*{Broader Impact}

The main contribution of the paper is a mathematically motivated attention mechanism which can be used for deep learning on point cloud based problems. We do not see a direct potential of \textit{negative} impact to the society. However, we would like to stress that this type of algorithm is inherently suited for classification and regression problems on molecules. The SE(3)-Transformer therefore lends itself to application in drug research. One concrete application we are currently investigating is to use the algorithm for early-stage suitability classification of molecules for inhibiting the reproductive cycle of the coronavirus. While research of this sort always requires intensive testing in wet labs, computer algorithms can be and are being used to filter out particularly promising compounds from large databases of millions of molecules.

%% file: appendix.tex
\newpage
\appendix

\section{Group Theory and Tensor Field Networks}
\paragraph{Groups}
A \emph{group} is an abstract mathematical concept. Formally a group $(G,\circ)$ consists of a set $G$ and a binary composition operator $\circ: G \times G \to G$ (typically we just use the symbol $G$ to refer to the group). All groups must adhere to the following 4 axioms
\begin{itemize}
    \item \textbf{Closure}: $g\circ h \in G$ for all $g, h \in G$
    \item \textbf{Associativity}: $f \circ (g \circ h) = (f \circ g) \circ h = f \circ g \circ h$ for all $f, g, h \in G$
    \item \textbf{Identity}: There exists an element $e \in G$ such that $e \circ g = g \circ e = g$ for all $g \in G$
    \item \textbf{Inverses}: For each $g \in G$ there exists a $g^{-1} \in G$ such that $g^{-1} \circ g = g \circ g^{-1} = e$
\end{itemize}
In practice, we omit writing the binary composition operator $\circ$, so would write $gh$ instead of $g \circ h$. Groups can be finite or infinite, countable or uncountable, compact or non-compact. Note that they are not necessarily \emph{commutative}; that is, $gh \neq hg$ in general.

\paragraph{Actions/Transformations}
Groups are useful concepts, because they allow us to describe the structure of \emph{transformations}, also sometimes called \emph{actions}. A transformation (operator) $T_g: \mc{X} \to \mc{X}$ is an injective map from a space into itself. It is parameterised by an element $g$ of a group $G$. Transformations obey two laws:
\begin{itemize}
    \item \textbf{Closure}: $T_g \circ T_h$ is a valid transformation for all $g, h \in G$
    \item \textbf{Identity}: There exists at least one element $e \in G$ such that $T_e[\bb{x}] = \bb{x}$ for all $\bb{x} \in \mc{X}$,
\end{itemize}
where $\circ$ denotes composition of transformations. For the expression $T_g[\bb{x}]$, we say that $T_g$ \emph{acts} on $\bb{x}$. It can also be shown that transformations are associative under composition. To codify the structure of a transformation, we note that due to closure we can always write
\begin{align}
    T_g \circ T_h = T_{gh},
\end{align}
If for any $x, y \in \mc{X}$ we can always find a group element $g$, such that $T_g[x] = y$, then we call $\mc{X}$ a \emph{homogeneous space}. Homogeneous spaces are important concepts, because to each pair of points $x, y$ we can always associate at least one group element.

\paragraph{Equivariance and Intertwiners}
As written in the main body of the text, equivariance is a property of functions $f: \mc{X} \to \mc{Y}$. Just to recap, given a set of transformations $T_g: \mc{X} \to \mc{X}$ for $g \in G$, where $G$ is an abstract group, a function $f: \mc{X} \to \mc{Y}$ is called equivariant if for every $g$ there exists a transformation $S_g: \mc{Y} \to \mc{Y}$ such that
\begin{align}
    S_g[f(\bb{x})] = f(T_g[\bb{x}]) \qquad \text{for all } g\in G, \bb{x}\in \mc{X}.
\end{align}
If $f$ is linear and equivariant, then it is called an intertwiner. Two important questions arise: 1) How do we choose $S_g$? 2) once we have $(T_g, S_g)$, how do we solve for $f$? To answer these questions, we need to understand what kinds of $S_g$ are possible. For this, we review representations.

\paragraph{Representations}
A group representation $\rho: G \to GL(N)$ is a map from a group $G$ to the set of $N\times N$ invertible matrices $GL(N)$. Critically $\rho$ is a \emph{group homomorphism}; that is, it satisfies the following property $\rho(g_1 g_2) = \rho(g_1)\rho(g_2)$ for all $g_1, g_2 \in G$. Representations can be used as transformation operators, acting on $N$-dimensional vectors $\bb{x} \in \mbb{R}^N$. For instance, for the group of 3D rotations, known as $SO(3)$, we have that 3D rotation matrices, $\rho(g) = \bb{R}_g$ act on (i.e., rotate) 3D vectors, as
\begin{align}
    T_g[\bb{x}] = \rho(g)\bb{x} = \bb{R}_g \bb{x}, \qquad \text{for all } \bb{x} \in \mc{X}, g \in G.
\end{align}
However, there are many more representations of $SO(3)$ than just the 3D rotation matrices. Among representations, two representations $\rho$ and $\rho'$ of the same dimensionality are said to be \emph{equivalent} if they can be connected by a similarity transformation
\begin{align}
    \rho'(g) = \bb{Q}^{-1} \rho(g) \bb{Q}, \qquad \text{for all } g \in G.
\end{align}
We also say that a representation is \emph{reducible} if is can be written as
\begin{align}
    \rho(g) = \bb{Q}^{-1} (\rho_1(g) \oplus \rho_2(g)) \bb{Q} = \bb{Q}^{-1} \begin{bmatrix} \rho_1(g) & \\ & \rho_2(g) \end{bmatrix} \bb{Q}, \qquad \text{for all } g \in G.
\end{align}
If the representations $\rho_1$ and $\rho_2$ are not reducible, then they are called \emph{irreducible representations} of $G$, or \emph{irreps}. In a sense, they are the atoms among representations, out of which all other representations can be constructed. Note that each irrep acts on a separate subspace, mapping vectors from that space back into it. We say that subspace $\mc{X}_\ell \in \mc{X}$ is \emph{invariant under} irrep $\rho_\ell$, if $\{\rho_\ell(g)\bb{x} \,\, | \,\, \bb{x} \in \mc{X}_\ell, g \in G \} \subseteq \mc{X}_\ell$.

\paragraph{Representation theory of $SO(3)$}
As it turns out, all linear representations of compact groups\footnote{Over a field of characteristic zero.} (such as $SO(3)$) can be decomposed into a direct sum of irreps, as
\begin{align}
    \rho(g) = \bb{Q}^{\top} \left [ \bigoplus_{J} \bb{D}_{J}(g) \right ] \bb{Q}, \label{eq:compact-decomposition}
\end{align}
where $\bb{Q}$ is an orthogonal, $N \times N$, change-of-basis matrix \citep{chirikjian2001engineering}; and each $\bb{D}_J$ for $J = 0,1,2,...$ is a $(2J+1) \times (2J + 1)$ matrix known as a Wigner-D matrix. The Wigner-D matrices are the irreducible representations of $SO(3)$. We also mentioned that vectors transforming according to $\bb{D}_J$ (i.e.\ we set $\bb{Q} = \bb{I}$), are called \emph{type}-$J$ vectors. Type-0 vectors are invariant under rotations and type-1 vectors rotate according to 3D rotation matrices. Note, type-$J$ vectors have length $2J+1$. In the previous paragraph we mentioned that irreps act on orthogonal subspaces $\mc{X}_0, \mc{X}_1, ...$. The orthogonal subspaces corresponding to the Wigner-D matrices are the space of \emph{spherical harmonics}.

\paragraph{The Spherical Harmonics}
The spherical harmonics $\bb{Y}_J: S^2 \to \mbb{C}^{2J+1}$ for $J \geq 0$ are square-integrable complex-valued functions on the sphere $S^2$. They have the satisfying property that they are rotated directly by the Wigner-D matrices as
\begin{align}
    \bb{Y}_J(\bb{R}_g^{-1}\bb{x}) = \bb{D}_{J}^*(g)\bb{Y}_{J}(\bb{x}), \qquad \bb{x} \in S^2, g\in G, \label{eq:spherical-harmonics}
\end{align}
where $\bb{D}_J$ is the $J$\textsuperscript{th} Wigner-D matrix and $\bb{D}_J^*$ is its complex conjugate. They form an orthonormal basis for (the Hilbert space of) square-integrable functions on the sphere $L^2(S^2)$, with inner product given as
\begin{align}
    \la f, h \ra_{S^2} = \int_{S^2} f(\bb{x}) h^*(\bb{x}) \, \mathrm d\bb{x}.
\end{align}
So $\la Y_{Jm}, Y_{J'm'} \ra_{S^2} = \delta_{JJ'} \delta_{mm'}$, where $Y_{Jm}$ is the $m$\textsuperscript{th} element of $\bb{Y}_J$. We can express any function in $L^2(S^2)$ as a linear combination of spherical harmonics, where
\begin{align}
    f(\bb{x}) = \sum_{J \geq 0} \bb{f}_J^\top \bb{Y}_{J}(\bb{x}), \qquad \bb{x} \in S^2,
\end{align}
where each $\bb{f}_J$ is a vector of coefficients of length $2J+1$. And in the opposite direction, we can retrieve the coefficients as
\begin{align}
    \bb{f}_J = \int_{S^2} f(\bb{x}) \bb{Y}_{J}^* (\bb{x}) \, \mathrm{d} \bb{x}
\end{align}
following from the orthonormality of the spherical harmonics.
This is in fact a Fourier transform on the sphere and the the vectors $\bb{f}_J$ can be considered Fourier coefficients. Critically, we can represent rotated functions as
\begin{align}
    f(\bb{R}_g^{-1}\bb{x}) = \sum_{J \geq 0} \bb{f}_J^\top \bb{D}_{J}^*(g)\bb{Y}_{J}(\bb{x}), \qquad \bb{x} \in S^2, g\in G.
\end{align}

\paragraph{The Clebsch-Gordan Decomposition}
In the main text we introduced the \emph{Clebsch-Gordan coefficients}. These are used in the construction of the equivariant kernels. They arise in the situation where we have a tensor product of Wigner-D matrices, which as we will see is part of the equivariance constraint on the form of the equivariant kernels. In representation theory a tensor product of representations is also a representation, but since it is not an easy object to work with, we seek to decompose it into a direct sum of irreps, which are easier. This decomposition is of the form of \Cref{eq:compact-decomposition}, written
\begin{align}
    \bb{D}_k(g) \otimes \bb{D}_\ell(g) = \bb{Q}^{\ell k\top} \left [ \bigoplus_{J = |k - \ell|}^{k + \ell} \bb{D}_J(g) \right ] \bb{Q}^{\ell k}.
\end{align}
In this specific instance, the change of basis matrices $\bb{Q}^{\ell k}$ are given the special name of the Clebsch-Gordan coefficients. These can be found in many mathematical physics libraries.

\paragraph{Tensor Field Layers}
In Tensor Field Networks \cite{ThomasSKYKR18} and 3D Steerable CNNs \cite{WeilerGWBC18}, the authors solve for the intertwiners between SO(3) equivariant point clouds. Here we run through the derivation again in our own notation.

We begin with a point cloud $f(\bb{x}) = \sum_{j=1}^N\bb{f}_j \delta(\bb{x} - \bb{x}_j)$, where $\bb{f}_j$ is an equivariant point feature. Let's say that $\bb{f}_j$ is a type-$k$ feature, which we write as $\bb{f}_j^k$ to remind ourselves of the fact. Now say we perform a convolution $*$ with kernel $\bb{W}^{\ell k}: \bb{R}^3 \to \mbb{R}^{(2\ell+1)\times(2k +1)}$, which maps from type-$k$ features to type-$\ell$ features. Then
\begin{align}
    \bb{f}_\text{out,i}^\ell &= [\bb{W}^{\ell k} * \bb{f}_{\text{in}}^k](\bb{x}) \\
    &= \int_{\mbb{R}^3} \bb{W}^{\ell k}(\bb{x}' - \bb{x}_i)\bb{f}_{\text{in}}^k(\bb{x}') \, \mathrm{d} \bb{x}' \\
    &= \int_{\mbb{R}^3} \bb{W}^{\ell k}(\bb{x}' - \bb{x}_i) \sum_{j=1}^N \bb{f}_{\text{in},j}^k\delta(\bb{x}' - \bb{x}_j) \, \mathrm{d} \bb{x}' \\
    &= \sum_{j=1}^N \int_{\mbb{R}^3} \bb{W}^{\ell k}(\bb{x}' - \bb{x}_i)  \bb{f}_{\text{in},j}^k\delta(\bb{x}' - \bb{x}_j) \, \mathrm{d} \bb{x}' && \text{change of variables } \bb{x}'' = \bb{x}' - \bb{x}_j \\
    &= \sum_{j=1}^N \int_{\mbb{R}^3} \bb{W}^{\ell k}(\bb{x}'' + \bb{x}_j - \bb{x}_i)  \bb{f}_{\text{in},j}^k\delta(\bb{x}'') \, \mathrm{d} \bb{x}'' && \text{sifting theorem }\\
    &= \sum_{j=1}^N \bb{W}^{\ell k}(\bb{x}_j - \bb{x}_i)  \bb{f}_{\text{in},j}^k. \label{eq:tfn-conv}
\end{align}
Now let's apply the equivariance condition to this expression, then
\begin{align}
    \bb{D}_\ell(g) \bb{f}_\text{out,i}^\ell &= \sum_{j=1}^N \bb{W}^{\ell k}(\bb{R}_g^{-1}(\bb{x}_j - \bb{x}_i)) \bb{D}_k(g) \bb{f}_{\text{in},j}^k \\
    \implies \bb{f}_\text{out,i}^\ell &= \sum_{j=1}^N \bb{D}_\ell(g)^{-1} \bb{W}^{\ell k}(\bb{R}_g^{-1}(\bb{x}_j - \bb{x}_i)) \bb{D}_k(g) \bb{f}_{\text{in},j}^k 
\end{align}
Now we notice that this expression should also be equal to \Cref{eq:tfn-conv}, which is the convolution with an unrotated point cloud. Thus we end up at
\begin{align}
\boxed{
    \bb{W}^{\ell k}(\bb{R}_g^{-1}\bb{x}) = \bb{D}_\ell(g) \bb{W}^{\ell k}(\bb{x}) \bb{D}_k(g)^{-1},
}
\end{align}
which is sometimes refered to as the \emph{kernel constraint}. To solve the kernel constraint, we notice that it is a linear equation and that we can rearrange it as
\begin{align}
    \text{vec}(\bb{W}^{\ell k}(\bb{R}_g^{-1}\bb{x})) = (\bb{D}_k(g) \otimes \bb{D}_\ell(g)) \text{vec}(\bb{W}^{\ell k}(\bb{x}))
\end{align}
where we used the identity $\text{vec}(\bb{AXB}) = (\bb{B}^\top \otimes \bb{A})\text{vec}(\bb{X})$ and the fact that the Wigner-D matrices are orthogonal. Using the Clebsch-Gordan decomposition we rewrite this as
\begin{align}
    \text{vec}(\bb{W}^{\ell k}(\bb{R}_g^{-1}\bb{x}))  = \bb{Q}^{\ell k\top} \left [ \bigoplus_{J = |k - \ell|}^{k + \ell} \bb{D}_J(g) \right ] \bb{Q}^{\ell k} \text{vec}(\bb{W}^{\ell k}(\bb{R}_g^{-1}\bb{x})) .
\end{align}
Lastly, we can left multiply both sides by $\bb{Q}^{\ell k}$ and denote $\bm{\eta}^{\ell k}(\bb{x}) \triangleq \bb{Q}^{\ell k} \text{vec}(\bb{W}^{\ell k}(\bb{x}))$, noting the the Clebsch-Gordan matrices are orthogonal. At the same time we 
\begin{align}
    \bm{\eta}^{\ell k}(\bb{R}_g^{-1}\bb{x}) =  \left [ \bigoplus_{J = |k - \ell|}^{k + \ell} \bb{D}_J(g) \right ] \bm{\eta}^{\ell k}(\bb{x}).
\end{align}
Thus we have that $\bm{\eta}^{\ell k}_J(\bb{R}_g^{-1}\bb{x})$ the $J$\textsuperscript{th} subvector of $\bm{\eta}^{\ell k}(\bb{R}_g^{-1}\bb{x})$ is subject to the constraint
\begin{align}
    \bm{\eta}_J^{\ell k}(\bb{R}_g^{-1}\bb{x}) =  \bb{D}_J(g) \bm{\eta}_J^{\ell k}(\bb{x}),
\end{align}
which is exactly the transformation law for the spherical harmonics from \Cref{eq:spherical-harmonics}! Thus one way how $\bb{W}^{\ell k}(\bb{x})$ can be constructed is
\begin{align}
    \text{vec} \left ( \bb{W}^{\ell k}(\bb{x}) \right ) = \bb{Q}^{\ell k \top} \bigoplus_{J = |k - \ell|}^{k + \ell} \bb{Y}_J(\bb{x}).
\end{align}

\section{Recipe for Building an Equivariant Weight Matrix}
One of the core operations in the SE(3)-Transformer is multiplying a feature vector $\bb{f}$, which transforms according to $SO(3)$, with a matrix $\bb{W}$ while preserving equivariance:
\begin{align}
S_g[\bb{W} * \bb{f}](\bb{x}) = [\bb{W} * T_g[\bb{f}]](\bb{x}),
\end{align}
where $T_g[\bb{f}](\bb{x}) = \rho_\text{in}(g)\bb{f}(\bb{R}_g^{-1}\bb{x})$ and $S_g[\bb{f}](\bb{x}) = \rho_\text{out}(g)\bb{f}(\bb{R}_g^{-1}\bb{x})$. Here, as in the previous section we showed how such a matrix $\bb{W}$ could be constructed when mapping between features of type-$k$ and type-$\ell$, where $\rho_\text{in}(g)$ is a block diagonal matrix of type-$k$ Wigner-D matrices and similarly $\rho_\text{in}(g)$ is made of type-$\ell$ Wigner-D matrices. $\bb{W}$ is dependent on the relative position $\bb{x}$ and underlies the linear equivariance constraints, but is also has learnable components, which we did not show in the previous section. In this section, we show how such a matrix is constructed in practice.

Previously we showed that 
\begin{align}
    \text{vec} \left ( \bb{W}^{\ell k}(\bb{x}) \right ) = \bb{Q}^{\ell k \top} \bigoplus_{J = |k - \ell|}^{k + \ell} \bb{Y}_J(\bb{x}),
\end{align}
which is an equivariant mapping between vectors of types $k$ and $\ell$. In practice, we have multiple input vectors $\{\bb{f}_c^k\}_c$ of type-$k$ and multiple output vectors of type-$\ell$. For simplicity, however, we ignore this and pretend we only have a single input and single output. Note that $\bb{W}^{\ell k}$ has no learnable components. Note that the kernel constraint only acts in the angular direction, but not in the radial direction, so we can introduce scalar radial functions $\varphi_{J}^{\ell k}: \bb{R}_{\geq 0} \to \bb{R}$ (one for each $J$), such that 
\begin{align}
    \text{vec} \left ( \bb{W}^{\ell k}(\bb{x}) \right ) = \bb{Q}^{\ell k \top} \bigoplus_{J = |k - \ell|}^{k + \ell} \varphi_J^{\ell k}(\|\bb{x}\|) \bb{Y}_J(\bb{x}),
\end{align}
The radial functions $\varphi_{J}^{\ell k}(\|\bb{x}\|)$ act as an independent, learnable scalar factor for each degree $J$. The vectorised matrix has dimensionality $(2\ell+1)(2k+1)$. We can unvectorise the above yielding 
\begin{align}
    \bb{W}^{\ell k}(\bb{x}) &= \text{unvec}\left(\bb{Q}^{\ell k \top} \bigoplus_{J = |k - \ell|}^{k + \ell} \varphi_J^{\ell k}(\|\bb{x}\|) \bb{Y}_J(\bb{x})\right) \\
    &= \sum_{J = |k - \ell|}^{k + \ell} \varphi_J^{\ell k}(\|\bb{x}\|) \text{unvec}\left( \bb{Q}_J^{\ell k \top}  \bb{Y}_J(\bb{x}) \right )
\end{align}
where $\bb{Q}_J^{\ell k}$ is a $(2\ell+1)(2k+1) \times (2J+1)$ slice from $\bb{Q}^{\ell k}$, corresponding to spherical harmonic $\bb{Y}_J.$ As we showed in the main text, we can also rewrite the unvectorised Clebsch-Gordan--spherical harmonic matrix-vector product as
\begin{align}
    \text{unvec}\left( \bb{Q}_J^{\ell k \top}  \bb{Y}_J(\bb{x}) \right ) = \sum_{m=-J}^J \bb{Q}_{Jm}^{\ell k \top}  Y_{Jm}(\bb{x}).
\end{align}
In contrast to \citet{WeilerGWBC18}, we do not voxelise space and therefore $\bb{x}$ will be different for each pair of points in each point cloud. However, the same $\bb{Y}_J\left(\bb{x}\right)$ will be used multiple times in the network and even multiple times in the same layer. Hence, precomputing them at the beginning of each forward pass for the entire network can significantly speed up the computation. The Clebsch-Gordan coefficients do not depend on the relative positions and can therefore be precomputed once and stored on disk. Multiple libraries exist which approximate those coefficients numerically.

\section{Accelerated Computation of Spherical Harmonics}
\label{sec:sh_accelerated}
\begin{figure}[t]
    \centering
    \begin{subfigure}[c]{0.45\linewidth}
        \centering
        \includegraphics[width=\linewidth]{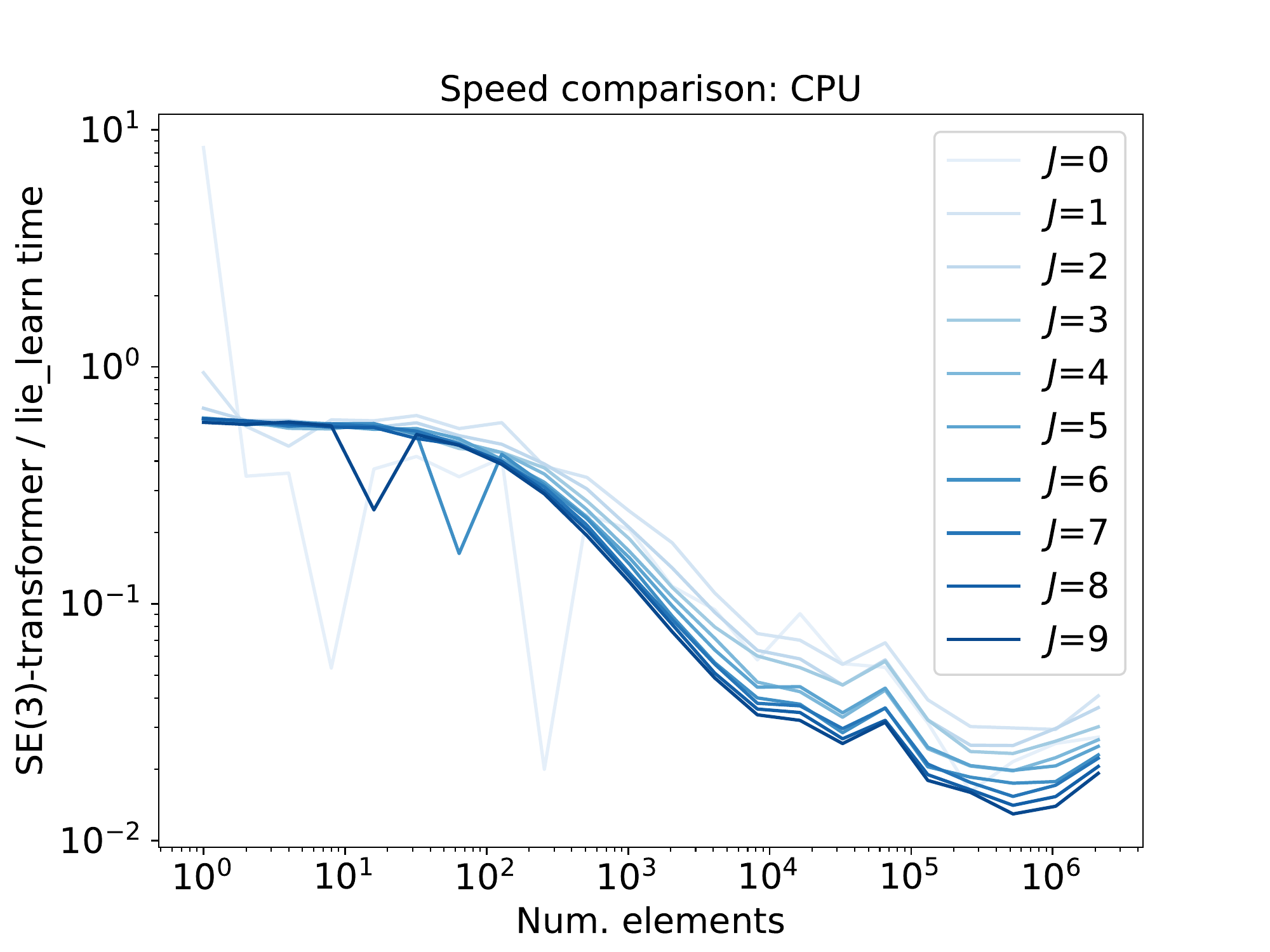}
        \subcaption{Speed comparison on the CPU.}
        \label{fig:speed_cpu}
    \end{subfigure}
    \hspace*{\fill}
    \begin{subfigure}[c]{0.45\linewidth}
        \centering
        \includegraphics[width=\linewidth]{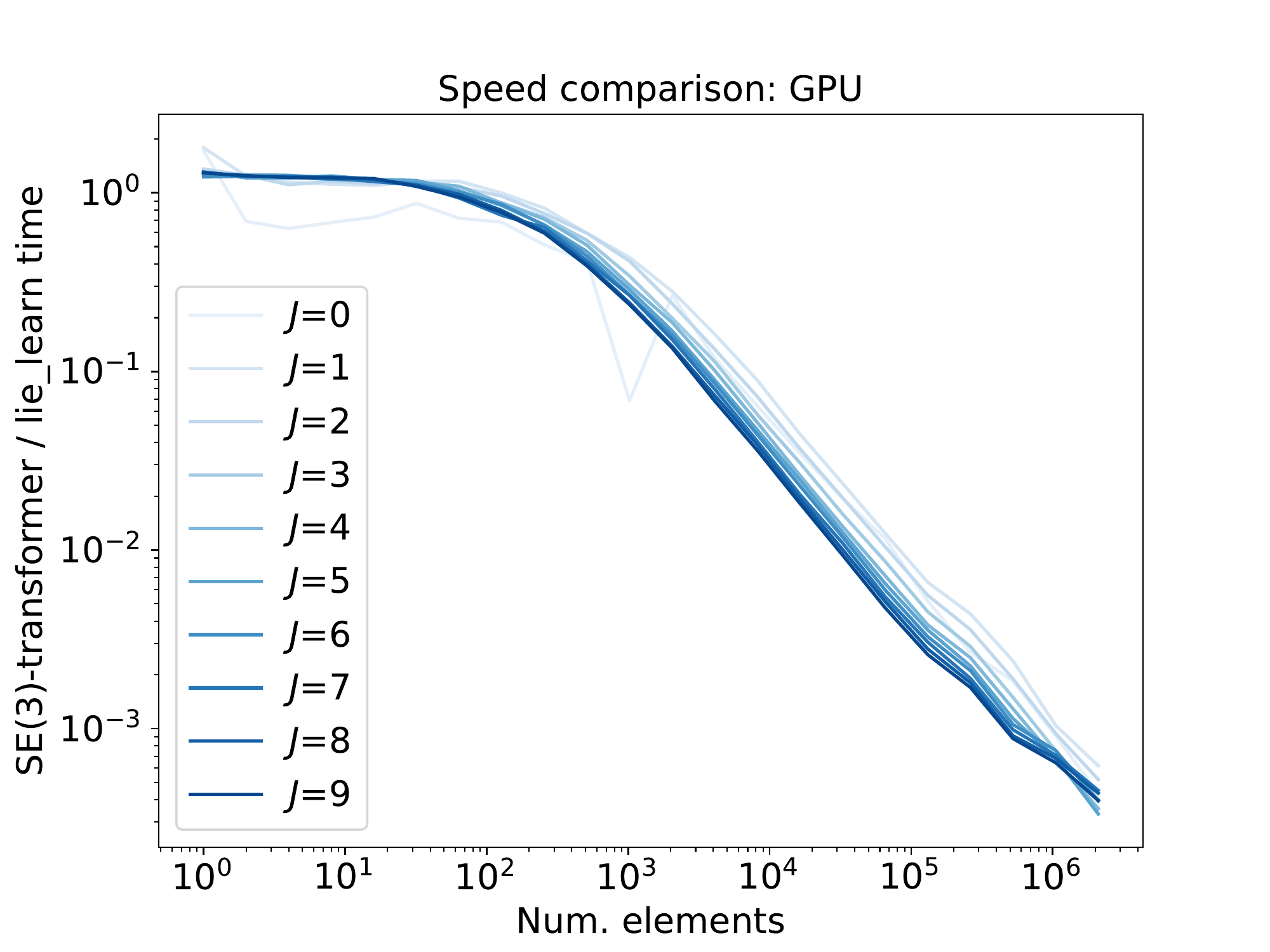}
        \subcaption{Speed comparison on the GPU.}
        \label{fig:speed_gpu}
    \end{subfigure}
    \caption{Spherical harmonics computation of our own implementation compared to the \texttt{lie-learn} library. We found that speeding up the computation of spherical harmonics is critical to scale up both Tensor Field Networks \citep{ThomasSKYKR18} and SE(3)-Transformers to solve real-world machine learning tasks.}
    \label{fig:speed_sh}
\end{figure}

The spherical harmonics (SH) typically have to be computed on the fly for point cloud methods based on irreducible computations, a bottleneck of TFNs \citep{ThomasSKYKR18}. \citet{ThomasSKYKR18} ameliorate this by restricting the maximum type of feature to type-2, trading expressivity for speed. \citet{WeilerGWBC18} circumvent this challenge by voxelising the input, allowing them to pre-compute spherical harmonics for fixed relative positions. This is at the cost of detail as well as exact rotation and translation equivariance.

The number of spherical harmonic lookups in a network based on irreducible representations can quickly become large (number of layers $\times$ number of points $\times$ number of neighbours $\times$ number of channels $\times$ number of degrees needed). This motivates parallelised computation on the GPU - a feature not supported by common libraries. To that end, we wrote our own spherical harmonics library in \texttt{Pytorch}, which can generate spherical harmonics on the GPU. We found this critical to being able to run the SE(3)-Transformer and Tensor Field network baselines in a reasonable time. This library is accurate to within machine precision against the \texttt{scipy} counterpart \texttt{scipy.special.sph\_harm} and is significantly faster.
E.g., for a ScanObjectNN model, we achieve $\sim 22\times$ speed up of the forward pass compared to a network built with SH from the \texttt{lielearn} library. A speed comparison isolating the computation of the spherical harmonics is shown in \Cref{fig:speed_sh}. Code is available at \url{https://github.com/FabianFuchsML/se3-transformer-public}. In the following, we outline our method to generate them. 

The tesseral/real spherical harmonics are given as
\begin{align}
    Y_{Jm}(\theta, \phi) = \sqrt{\frac{2 J + 1}{4 \pi}\frac{(J-m)!}{(J+m)!}} P_J^{|m|}(\cos \theta) \cdot \left \{ \begin{matrix*}[l]
     \sin(|m| \phi)  & m < 0,\\
    1 & m = 0, \\
    \cos(m \phi) & m > 0,
    \end{matrix*} \right .
\end{align}
where $P_J^{|m|}$ is the associated Legendre polynomial (ALP), $\theta \in [0, 2\pi)$ is azimuth, and $\phi \in [0,\pi]$ is a polar angle. The term $P_J^{|m|}$ is by far the most expensive component to compute and can be computed recursively. To speed up the computation, we use dynamic programming storing intermediate results in a memoization. 

We make use of the following recursion relations in the computation of the ALPs:
\begin{align}
    P_J^{|J|}(x) &= (-1)^{|J|} \cdot (1-x^2)^{|J|/2} \cdot (2|J|-1)!! && \text{boundary: $J=m$} \label{eq:leg-boundary} \\
    P_J^{-m}(x) &= (-1)^J \frac{(\ell - m)!}{(\ell + m)!} P_J^{m}(x) && \text{negate $m$} \label{eq:leg-negate}\\
    P_J^{|m|}(x) &= \frac{2J-1}{J-|m|} x P_{J-1}^{m}(x) + \mbb{I}[J - |m| > 1] \frac{J + |m| - 1}{J - |m|} P_{J-2}^{m}(x) && \text{recurse} \label{eq:leg-recurse}
\end{align}
where the \emph{semifactorial} is defined as $x!! = x (x-2) (x-4) \cdots$, and $\mbb{I}$ is the indicator function. These relations are helpful because they define a recursion. 

\begin{figure}
    \centering
    \includegraphics{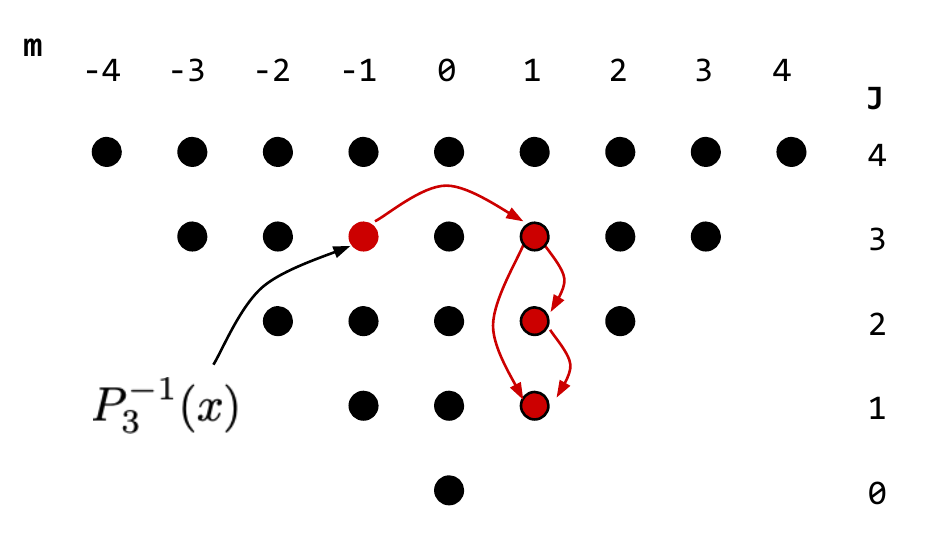}
    \caption{Subproblem graph for the computatin of the associated Legendre polynomials. To compute $P_3^{-1}(x)$, we compute $P_3^{1}(x)$, for which we need $P_2^{1}(x)$ and $P_1^{1}(x)$. We store each intermediate computation, speeding up average computation time by a factor of $\sim10$ on CPU.}
    \label{fig:subproblem}
\end{figure}

To understand how we recurse, we consider an example. \Cref{fig:subproblem} shows the space of $J$ and $m$. The black vertices represent a particular ALP, for instance, we have highlighted $P_3^{-1}(x)$. When $m <0$, we can use \Cref{eq:leg-negate} to compute $P_3^{-1}(x)$ from $P_3^1(x)$. We can then use 
\Cref{eq:leg-recurse} to compute $P_3^1(x)$ from $P_2^1(x)$ and $P_1^1(x)$. $P_2^1(x)$ can also be computed from \Cref{eq:leg-recurse} and the boundary value $P_1^1(x)$ can be computed directly using \Cref{eq:leg-boundary}. Crucially, all intermediate ALPs are stored for reuse. Say we wanted to compute $P_4^{-1}(x)$, then we could use \Cref{eq:leg-negate} to find it from $P_4^{-1}(x)$, which can be recursed from the stored values $P_3^1(x)$ and $P_2^1(x)$, without needing to recurse down to the boundary.

\section{Experimental Details}
\subsection{ScanObjectNN}
\label{as}
\subsubsection{SE(3)-Transformer and Tensor Field Network}
A particularity of object classification from point clouds is the large number of points the algorithm needs to handle. We use up to 200 points out of the available 2024 points per sample and create neighbourhoods with up to 40 nearest neighbours. It is worth pointing out that especially in this setting, adding self-attention (i.e. when comparing the SE(3) Transformer to Tensor Field Networks) significantly increased the stability. As a result, whenever we swapped out the attention mechanism for a convolution to retrieve the Tensor Field network baseline, we had to decrease the model size to obtain stable training. However, we would like to stress that all the Tensor Field networks we trained were significantly bigger than in the original paper \cite{ThomasSKYKR18}, mostly enabled by the faster computation of the spherical harmonics.

For the ablation study in \Cref{fig:scanobject_rotate}, we trained networks with 4 hidden equivariant layers with 5 channels each, and up to representation degree 2. This results in a hidden feature size per point of
\begin{align}
    5\cdot \sum_{\ell=0}^2 (2\ell + 1) = 45
\end{align}
We used 200 points of the point cloud and neighbourhood size 40. For the Tensor Field network baseline, in order to achieve stable training, we used a smaller model with 3 instead of 5 channels, 100 input points and neighbourhood size 10, but with representation degrees up to 3.

We used 1 head per attention mechanism yielding one attention weight for each pair of points but across all channels and degrees (for an implementation of multi-head attention, see \Cref{sec:QM9_details}). For the query embedding, we used the identity matrix. For the key embedding, we used a square equivariant matrix preserving the number of degrees and channels per degree.

For the quantitative comparison to the start-of-the-art in \Cref{tab:Scan}, we used 128 input points and neighbourhood size 10 for both the Tensor Field network baseline and the SE(3)-Transformer. We used farthest point sampling with a random starting point to retrieve the 128 points from the overall point cloud. We used degrees up to 3 and 5 channels per degree, which we again had to reduce to 3 channels  for the Tensor Field network to obtain stable training. We used a norm based non-linearity for the Tensor Field network (as in \cite{ThomasSKYKR18}) and no extra non-linearity (beyond the \textit{softmax} in the self-attention algorithm) for the SE(3) Transformer.

For all experiments, the final layer of the equivariant encoder maps to 64 channels of degree 0 representations. This yields a 64-dimensional SE(3) \textit{invariant} representation per point. Next, we pool over the point dimension followed by an MLP with one hidden layer of dimension 64, a ReLU and a 15 dimensional output with a cross entropy loss. We trained for 60000 steps with batch size 10. We used the Adam optimizer \cite{KingmaB14} with a start learning of \texttt{1e-2} and a reduction of the learning rate by 70\% every 15000 steps. Training took up to 2 days on a system with 4 CPU cores, 30 GB of RAM, and an NVIDIA GeForce GTX 1080 Ti GPU.

The input to the Tensorfield network and the Se(3) Transformer are relative x-y-z positions of each point w.r.t. their neighbours. To guarantee equivariance, these inputs are provided as fields of degree 1. For the `+z` versions, however, we deliberately break the SE(3) equivariance by providing additional and relative z-position as two additional scalar fields (i.e. degree 0), as well as relative x-y positions as a degree 1 field (where the z-component is set to 0).

\subsubsection{Number of Input Points}
\label{sec:app_n_points}
Limiting the input to 128/200 points in our experiments on ScanObjectNN was not primarily due to computational limitations: we conducted experiments with up to 2048 points, but without performance improvements. We suspect this is due to the global pooling. Examining cascaded pooling via attention is a future research direction. Interestingly, when limiting other methods to using 128 points, the SE(3)-Transformer outperforms the baselines (PointCNN: $80.3\pm 0.8\%$, PointGLR: $81.5 \pm 1.0\%$, DGCNN: $82.2\pm0.8\%$, \textbf{ours}: $85.0\pm 0.7\%$). It is worth noting that these methods were explicitly designed for the well-studied task of point classification whereas the SE(3)-Transformer was applied as is. Combining different elements from the current state-of-the-art methods with the geometrical inductive bias of the SE(3)-Transformer could potentially yield additional performance gains, especially with respect to leveraging inputs with more points. It is also worth noting that the SE(3)-Transformer was remarkably stable with respect to lowering the number of input points in an ablation study (16 points: $79.2\%$, 32 points: $81.4\%$, 64 points: $82.5\%$, 128 points: $85.0\%$, 256 points: $82.6\%$).

\subsubsection{Sample Complexity}
Equivariance is known to often lead to smaller sample complexity, meaning that less training data is needed (Fig. 10 in \citet{WorrallGRB2016}, Fig. 4 in \citet{WinkelsCohen}, Fig. 4 in \citet{WeilerGWBC18}). We conducted experiments with different amounts of training samples $N_{\text{samples}}$ from the ScanObjectNN dataset. The results showed that for all $N_{\text{samples}}$, the SE(3)-Transformer outperformed the Set Transformer, a non-equivariant network based on attention. The performance delta was also slightly higher for the smallest $N_{\text{samples}}$ we tested ($3.1\%$ of the samples available in the training split of ScanObjectNN) than when using all the data indicating that the SE(3)-Transformer performs particularly well on small amounts of training data. However, performance differences can be due to multiple reasons. Especially for small datasets, such as ScanObjectNN, where the performance does not saturate with respect to amount of data available, it is difficult to draw conclusions about sample complexity of one model versus another. In summary, we found that our experimental results are in line with the claim that equivariance decreases sample complexity in this specific case but do not give definitive support.

\subsubsection{Baselines}
\paragraph{DeepSet Baseline} We originally replicated the implementation proposed in \cite{Zaheer2017} for their object classification experiment on ModelNet40 \citep{modelnet40}. However, most likely due to the relatively small number of objects in the ScanObjectNN dataset, we found that reducing the model size helped the performance significantly. The reported model had 128 units per hidden layer (instead of 256) and no dropout but the same number of layers and type of non-linearity as in \cite{Zaheer2017}.

\paragraph{Set Transformer Baseline} We used the same architecture as \cite{SetTransformer} in their object classification experiment on ModelNet40 \citep{modelnet40} with an ISAB (induced set attention block)-based encoder followed by PMA (pooling by multihead attention) and an MLP.

\subsection{Relational Inference}
Following \citet{KipfFWWZ18}, we simulated trajectories for \(5\) charged, interacting particles. Instead of a 2d simulation setup, we considered a 3d setup. 
Positive and negative charges were drawn as Bernoulli trials (\(p=0.5\)).
We used the provided code base \texttt{https://github.com/ethanfetaya/nri} with the following modifications:
While we randomly sampled initial positions inside a \([-5,5]^3\) box, we removed the bounding-boxes during the simulation. We generated \(5\text{k}\) simulation samples for training and \(1\text{k}\) for testing. Instead of phrasing it as a time-series task, we posed it as a regression task: The input data is positions and velocities at a random time step as well as the signs of the charges. The labels (which the algorithm is learning to predict) are the positions and velocities 500 simulation time steps into the future.

\paragraph{Training Details}
We trained each model for 100,000 steps with batch size 128 using an Adam optimizer \cite{KingmaB14}. We used a fixed learning rate throughout training and conducted a separate hyper parameter search for each model to find a suitable learning rate.

\paragraph{SE(3)-Transformer Architecture}
We trained an SE(3)-Transformer with 4 equivariant layers, where the hidden layers had representation degrees $\{0,1,2,3\}$ and 3 channels per degree. The input is handled as two type-1 fields (for positions and velocities) and one type-0 field (for charges). The learning rate was set to \texttt{3e-3}. Each layer included attentive self-interaction.

We used 1 head per attention mechanism yielding one attention weight for each pair of points but across all channels and degrees (for an implementation of multi-head attention, see \Cref{sec:QM9_details}). For the query embedding, we used the identity matrix. For the key embedding, we used a square equivariant matrix preserving the number of degrees and channels per degree.

\paragraph{Baseline Architectures}
All our baselines fulfill permutation invariance (ordering of input points), but only the Tensor Field network and the linear baseline are SE(3) equivariant. For the \textbf{Tensor Field Network}\cite{ThomasSKYKR18} baseline, we used the same hyper parameters as for the SE(3) Transformer but with a linear self-interaction and an additional norm-based nonlinearity in each layer as in \citet{ThomasSKYKR18}. For the \textbf{DeepSet}\cite{Zaheer2017} baseline, we used 3 fully connected layers, a pooling layer, and two more fully connected layers with 64 units each. All fully connected layers act pointwise. The pooling layer uses max pooling to aggregate information from all points, but concatenates this with a skip connection for each point. Each hidden layer was followed by a LeakyReLU. The learning rate was set to \texttt{1e-3}. For the \textbf{Set Transformer}\cite{SetTransformer}, we used 4 self-attention blocks with 64 hidden units and 4 heads each. For each point this was then followed by a fully connected layer (64 units), a LeakyReLU and another fully connected layer. The learning rate was set to \texttt{3e-4}.

For the \textbf{linear baseline}, we simply propagated the particles linearly according to the simulation hyperparamaters. The linear baseline can be seen as removing the interactions between particles from the prediction. Any performance improvement beyond the linear baseline can therefore be interpreted as an indication for relational reasoning being performed.

\subsection{QM9} \label{sec:QM9_details} The QM9 regression dataset \citep{RamakrishnanDRvL14} is a publicly available chemical property prediction task consisting of 134k small drug-like organic molecules with up to 29 atoms per molecule. There are 5 atomic species (Hydrogen, Carbon, Oxygen, Nitrogen, and Flourine) in a molecular graph connected by chemical bonds of 4 types (single, double, triple, and aromatic bonds). `Positions' of each atom, measured in {\aa}ngtr\"oms, are provided. We used the exact same train/validation/test splits as \citet{anderson2019cormorant} of sizes 100k/18k/13k. 

\begin{figure}[h]
    \centering
    \includegraphics[width=0.5\textwidth]{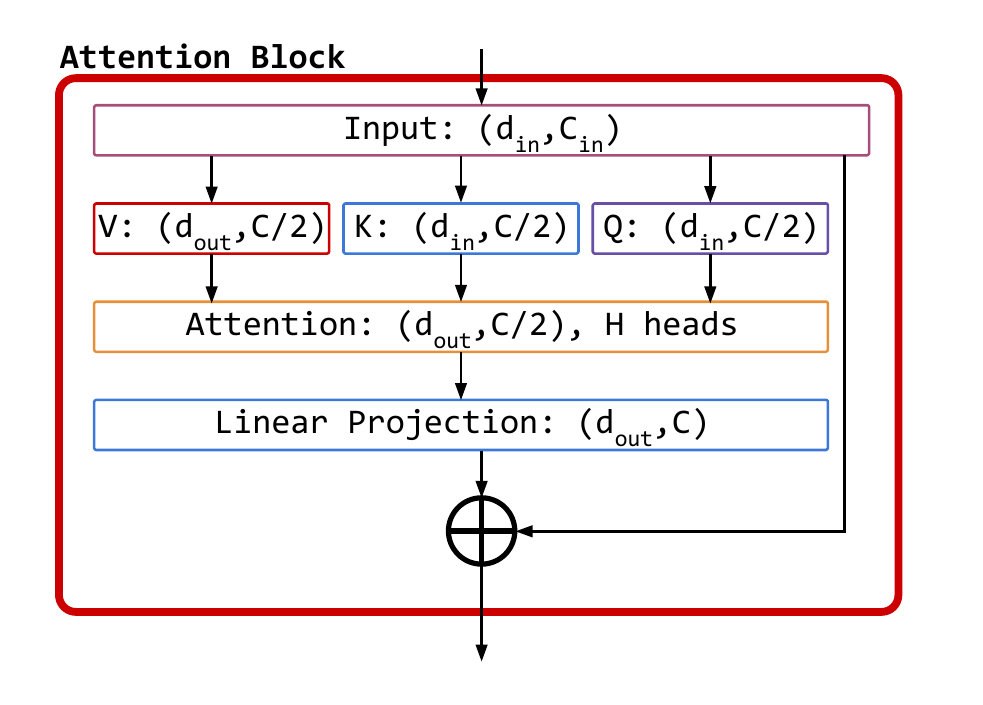}
    \caption{Attention block for the QM9 dataset. Each component is listed with a tuple of numbers representing the output feature types and multiplicities, so $(4,32)$ means feature types $0,1,2,3$ (with dimensionalities $1,3,5,7$), with 32 channels per type. }
    \label{fig:mab_qm9}
\end{figure}

The architecture we used is shown in Table \ref{tab:qm9_architecture}. It consists of 7 multihead attention layers interspersed with norm nonlinearities, followed by a TFN layer, max pooling, and two linear layers separated by a ReLU. For each attention layer, shown in \Cref{fig:mab_qm9}, we embed the input to half the number of feature channels before applying multiheaded attention \cite{VaswaniSPUJGKP17}. Multiheaded attention is a variation of attention, where we partition the queries, keys, and values into $H$ \emph{attention heads}. So if our embeddings have dimensionality $(4,16)$ (denoting 4 feature types with 16 channels each) and we use $H=8$ attention heads, then we partition the embeddings to shape $(4,2)$. We then combine each of the 8 sets of shape $(4,2)$ queries, keys, and values individually and then concatenate the results into a single vector of the original shape $(4,16)$. The keys and queries are edge embeddings, and thus the embedding matrices are of TFN type (c.f.\ \Cref{eq:equivariant-kernel}). For TFN type layers, the radial functions are learnable maps. For these we used neural networks with architecture shown in \Cref{tab:qm9_radial}.

For the norm nonlinearities \cite{WorrallGRB2016}, we use 
\begin{align}
\label{eq:norm-nonlinearity}
    \texttt{Norm ReLU}(\bb{f}^{\ell}) = \texttt{ReLU}(\texttt{LN} \left ( \|\bb{f}^{\ell}\| \right ) ) \cdot \frac{\bb{f}^{\ell}}{\|\bb{f}^{\ell}\|}, \qquad \text{where }
    \|\bb{f}^\ell\| = \sqrt {\sum_{m=-\ell}^\ell (f_m^{\ell})^2 },
\end{align}
where \texttt{LN} is layer norm \cite{BaKH16} applied across all features within a feature type. For the TFN baseline, we used the exact same architecture but we replaced each of the multiheaded attention blocks with a TFN layer with the same output shape.

The input to the network is a sparse molecular graph, with edges represented by the molecular bonds. The node embeedings are a 6 dimensional vector composed of a 5 dimensional one-hot embedding of the 5 atomic species and a 1 dimension integer node embedding for number of protons per atom. The edges embeddings are a 5 dimensional vector consisting of a 4 dimensional one-hot embedding of bond type and a positive scalar for the Euclidean distance between the two atoms at the ends of the bond. For each regression target, we normalised the values by mean and dividing by the standard deviation of the training set.

We trained for 50 epochs using Adam \cite{KingmaB14} at initial learning rate \texttt{1e-3} and a single-cycle cosine rate-decay to learning rate \texttt{1e-4}. The batch size was 32, but for the TFN baseline we used batch size 16, to fit the model in memory. We show results on the 6 regression tasks not requiring thermochemical energy subtraction in \Cref{tab:QM9}. As is common practice, we optimised architectures and hyperparameters on $\varepsilon_\text{HOMO}$ and retrained each network on the other tasks. Training took about 2.5 days on an NVIDIA GeForce GTX 1080 Ti GPU with 4 CPU cores and 15 GB of RAM. 

\begin{table}[h]
    \centering
    \caption{QM9 Network architecture: $d_\text{out}$ is the number of feature types of degrees $0,1,...,d_\text{out}-1$ at the output of the corresponding layer and $C$ is the number of multiplicities/channels per feature type. For the norm nonlinearity we use ReLUs, preceded by equivariant layer norm \cite{WeilerGWBC18} with learnable affine transform.}
    \label{tab:qm9_architecture}
    \begin{tabular}{cllll}
        \toprule
        \textsc{No. repeats} & \textsc{Layer Type} & $d_\text{out}$ & $C$\\
        1x & Input & 1 & 6 \\
        \midrule
        \multirow{2}{*}{1x}& Attention: 8 heads & 4 & 16 \\
        &Norm Nonlinearity & 4 & 16 \\
        \midrule
        \multirow{2}{*}{6x}& Attention: 8 heads & 4 & 16 \\
        &Norm Nonlinearity & 4 & 16 \\
        \midrule
        1x & TFN layer & 1 & 128 \\
        \midrule
        1x   & Max pool & 1 & 128 \\
        1x   & Linear &1 & 128\\
        1x   & ReLU & 1 & 128\\
        1x   & Linear &1 & 1 \\
        \bottomrule
    \end{tabular}
\end{table}

\begin{table}[h]
    \centering
    \caption{QM9 Radial Function Architecture. $C$ is the number of output channels at each layer. Layer norm \cite{BaKH16} is computed per pair of input and output feature types, which have $C_\text{in}$ and $C_\text{out}$ channels each.}
    \label{tab:qm9_radial}
    \begin{tabular}{llll}
        \toprule
        \textsc{Layer Type} & $C$\\
        \midrule
        Input & 6 \\
        \midrule
        Linear & 32 \\
        Layer Norm & 32 \\
        ReLU    & 32 \\
        \midrule
        Linear & 32 \\
        Layer Norm & 32 \\
        ReLU    & 32 \\
        \midrule
        Linear & $d_\text{out}*C_\text{in}*C_\text{out}$ \\
        \bottomrule
    \end{tabular}
\end{table}

\subsection{General Remark}
Across experiments on different datasets with the SE(3)-Transformer, we made the observation that the number of representation degrees have a significant but saturating impact on performance. A big improvement was observed when switching from degrees $\{0,1\}$ to $\{0,1,2\}$. Adding type-3 latent representations gave small improvements, further representation degrees did not change the performance of the model. However, higher representation degrees have a significant impact on memory usage and computation time. We therefore recommend representation degrees up to 2, when computation time and memory usage is a concern, and 3 otherwise.

%% file: EA4PC.bbl
\begin{thebibliography}{47}
\providecommand{\natexlab}[1]{#1}
\providecommand{\url}[1]{\texttt{#1}}
\expandafter\ifx\csname urlstyle\endcsname\relax
  \providecommand{\doi}[1]{doi: #1}\else
  \providecommand{\doi}{doi: \begingroup \urlstyle{rm}\Url}\fi

\bibitem[Anderson et~al.(2019)Anderson, Hy, and Kondor]{anderson2019cormorant}
Brandon Anderson, Truong~Son Hy, and Risi Kondor.
\newblock Cormorant: Covariant molecular neural networks.
\newblock In \emph{Advances in Neural Information Processing Systems
  (NeurIPS)}. 2019.

\bibitem[Ba et~al.(2016)Ba, Kiros, and Hinton]{BaKH16}
Lei~Jimmy Ba, Jamie~Ryan Kiros, and Geoffrey~E. Hinton.
\newblock Layer normalization.
\newblock \emph{arXiv Preprint}, 2016.

\bibitem[Ben-Shabat et~al.(2018)Ben-Shabat, Lindenbaum, and Fischer.]{3dmfv}
Yizhak Ben-Shabat, Michael Lindenbaum, and Anath Fischer.
\newblock 3dmfv: Three-dimensional point cloud classification in realtime using
  convolutional neural networks.
\newblock \emph{IEEE Robotics and Automation Letters}, 2018.

\bibitem[Chen et~al.(2019)Chen, Li, Xu, Chen, Wang, and Lin]{ClusterNet}
Chao Chen, Guanbin Li, Ruijia Xu, Tianshui Chen, Meng Wang, and Liang Lin.
\newblock {ClusterNet: Deep Hierarchical Cluster Network with Rigorously
  Rotation-Invariant Representation for Point Cloud Analysis}.
\newblock In \emph{Proceedings of the IEEE Conference on Computer Vision and
  Pattern Recognition (CVPR)}, 2019.

\bibitem[Chirikjian et~al.(2001)Chirikjian, Kyatkin, and
  Buckingham]{chirikjian2001engineering}
Gregory~S Chirikjian, Alexander~B Kyatkin, and AC~Buckingham.
\newblock Engineering applications of noncommutative harmonic analysis: with
  emphasis on rotation and motion groups.
\newblock \emph{Appl. Mech. Rev.}, 54\penalty0 (6):\penalty0 B97--B98, 2001.

\bibitem[Cohen and Welling(2017)]{CohenW2016}
Taco~S. Cohen and Max Welling.
\newblock Steerable cnns.
\newblock \emph{International Conference on Learning Representations (ICLR)},
  2017.

\bibitem[Cohen et~al.(2018)Cohen, Geiger, Koehler, and
  Welling]{cohen2018spherical}
Taco~S. Cohen, Mario Geiger, Jonas Koehler, and Max Welling.
\newblock Spherical cnns.
\newblock In \emph{International Conference on Learning Representations
  (ICLR)}, 2018.

\bibitem[Finzi et~al.(2020)Finzi, Stanton, Izmailov, and Wilson]{Finzi2020}
Marc Finzi, Samuel Stanton, Pavel Izmailov, and Andrew Wilson.
\newblock Generalizing convolutional neural networks for equivariance to lie
  groups on arbitrary continuous data.
\newblock \emph{Proceedings of the International Conference on Machine
  Learning, {ICML}}, 2020.

\bibitem[Fuchs et~al.(2020)Fuchs, Kosiorek, Sun, Jones, and Posner]{Mohart}
Fabian~B. Fuchs, Adam~R. Kosiorek, Li~Sun, Oiwi~Parker Jones, and Ingmar
  Posner.
\newblock End-to-end recurrent multi-object tracking and prediction with
  relational reasoning.
\newblock \emph{arXiv preprint}, 2020.

\bibitem[Gilmer et~al.(2020)Gilmer, Schoenholz, Riley, and Dahl]{Gilmer2017}
Justin Gilmer, Samuel~S. Schoenholz, Patrick~F. Riley, and Oriol Vinyals
  George~E. Dahl.
\newblock Neural message passing for quantum chemistry.
\newblock \emph{Proceedings of the International Conference on Machine
  Learning, {ICML}}, 2020.

\bibitem[Hirn et~al.(2017)Hirn, Mallat, and Poilvert]{HirnMP17}
Matthew~J. Hirn, St{\'{e}}phane Mallat, and Nicolas Poilvert.
\newblock Wavelet scattering regression of quantum chemical energies.
\newblock \emph{Multiscale Model. Simul.}, 15\penalty0 (2):\penalty0 827--863,
  2017.

\bibitem[Hoshen(2017)]{hoshen2017vain}
Yedid Hoshen.
\newblock Vain: Attentional multi-agent predictive modeling.
\newblock \emph{Advances in Neural Information Processing Systems (NeurIPS)},
  2017.

\bibitem[Kingma and Ba(2015)]{KingmaB14}
Diederik~P. Kingma and Jimmy Ba.
\newblock Adam: {A} method for stochastic optimization.
\newblock In \emph{International Conference on Learning Representations,
  {ICLR}}, 2015.

\bibitem[Kipf et~al.(2018)Kipf, Fetaya, Wang, Welling, and Zemel]{KipfFWWZ18}
Thomas~N. Kipf, Ethan Fetaya, Kuan{-}Chieh Wang, Max Welling, and Richard~S.
  Zemel.
\newblock Neural relational inference for interacting systems.
\newblock In \emph{Proceedings of the International Conference on Machine
  Learning, {ICML}}, 2018.

\bibitem[Kondor(2018)]{kondor2018nbody}
Risi Kondor.
\newblock N-body networks: a covariant hierarchical neural network architecture
  for learning atomic potentials.
\newblock \emph{arXiv preprint}, 2018.

\bibitem[Lee et~al.(2019)Lee, Lee, Kim, Kosiorek, Choi, and
  Teh]{SetTransformer}
Juho Lee, Yoonho Lee, Jungtaek Kim, Adam~R. Kosiorek, Seungjin Choi, and
  Yee~Whye Teh.
\newblock Set transformer: {A} framework for attention-based
  permutation-invariant neural networks.
\newblock In \emph{Proceedings of the International Conference on Machine
  Learning, {ICML}}, 2019.

\bibitem[Lin et~al.(2017)Lin, Feng, dos Santos, Yu, Xiang, Zhou, and
  Bengio]{lin2017structured}
Zhouhan Lin, Minwei Feng, Cicero~Nogueira dos Santos, Mo~Yu, Bing Xiang, Bowen
  Zhou, and Yoshua Bengio.
\newblock A structured self-attentive sentence embedding.
\newblock \emph{International Conference on Learning Representations (ICLR)},
  2017.

\bibitem[Parmar et~al.(2019)Parmar, Ramachandran, Vaswani, Bello, Levskaya, and
  Shlens]{ParmarRVBLS19}
Niki Parmar, Prajit Ramachandran, Ashish Vaswani, Irwan Bello, Anselm Levskaya,
  and Jon Shlens.
\newblock Stand-alone self-attention in vision models.
\newblock In \emph{Advances in Neural Information Processing System (NeurIPS)},
  2019.

\bibitem[Qi et~al.(2017{\natexlab{a}})Qi, Su, Mo, and Guibas]{Pointnet}
Charles~R Qi, Hao Su, Kaichun Mo, and Leonidas~J Guibas.
\newblock Pointnet: Deep learning on point sets for 3d classification and
  segmentation.
\newblock \emph{IEEE Conference on Computer Vision and Pattern Recognition
  (CVPR)}, 2017{\natexlab{a}}.

\bibitem[Qi et~al.(2017{\natexlab{b}})Qi, Yi, Su, and Guibas]{Pointnetpp}
Charles~R Qi, Li~Yi, Hao Su, and Leonidas~J Guibas.
\newblock Pointnet++: Deep hierarchical feature learning on point sets in a
  metric space.
\newblock \emph{Advances in Neural Information Processing Systems (NeurIPS)},
  2017{\natexlab{b}}.

\bibitem[Ramakrishnan et~al.(2014)Ramakrishnan, Dral, Rupp, and von
  Lilienfeld]{RamakrishnanDRvL14}
Raghunathan Ramakrishnan, Pavlo Dral, Matthias Rupp, and Anatole von
  Lilienfeld.
\newblock Quantum chemistry structures and properties of 134 kilo molecules.
\newblock \emph{Scientific Data}, 1, 08 2014.

\bibitem[Rao et~al.(2019)Rao, Lu, and Zhou]{RaoFractal}
Yongming Rao, Jiwen Lu, and Jie Zhou.
\newblock Spherical fractal convolutional neural networks for point cloud
  recognition.
\newblock In \emph{Proceedings of the IEEE Conference on Computer Vision and
  Pattern Recognition (CVPR)}, 2019.

\bibitem[Rao et~al.(2020)Rao, Lu, and Zhou]{PointGLR}
Yongming Rao, Jiwen Lu, and Jie Zhou.
\newblock Global-local bidirectional reasoning for unsupervised representation
  learning of 3d point clouds.
\newblock In \emph{Proceedings of the IEEE Conference on Computer Vision and
  Pattern Recognition (CVPR)}, 2020.

\bibitem[Romero et~al.(2020)Romero, Bekkers, Tomczak, and
  Hoogendoorn]{Romero2020}
David~W. Romero, Erik~J. Bekkers, Jakub~M. Tomczak, and Mark Hoogendoorn.
\newblock Attentive group equivariant convolutional networks.
\newblock \emph{Proceedings of the International Conference on Machine Learning
  (ICML)}, 2020.

\bibitem[Schütt et~al.(2017)Schütt, Kindermans, Sauceda, Chmiela1,
  Tkatchenko, and Müller]{schnet}
K.~T. Schütt, P.-J. Kindermans, H.~E. Sauceda, S.~Chmiela1, A.~Tkatchenko, and
  K.-R. Müller.
\newblock Schnet: A continuous-filter convolutional neural network for modeling
  quantum interactions.
\newblock \emph{Advances in Neural Information Processing Systems (NeurIPS)},
  2017.

\bibitem[Shaw et~al.(2018)Shaw, Uszkoreit, and Vaswani]{shaw2018selfattention}
Peter Shaw, Jakob Uszkoreit, and Ashish Vaswani.
\newblock Self-attention with relative position representations.
\newblock \emph{Annual Conference of the North American Chapter of the
  Association for Computational Linguistics (NAACL-HLT)}, 2018.

\bibitem[Sosnovik et~al.(2020)Sosnovik, Szmaja, and Smeulders]{Sosnovik20}
Ivan Sosnovik, Michał Szmaja, and Arnold Smeulders.
\newblock Scale-equivariant steerable networks.
\newblock \emph{International Conference on Learning Representations (ICLR)},
  2020.

\bibitem[Thomas et~al.(2018)Thomas, Smidt, Kearnes, Yang, Li, Kohlhoff, and
  Riley]{ThomasSKYKR18}
Nathaniel Thomas, Tess Smidt, Steven~M. Kearnes, Lusann Yang, Li~Li, Kai
  Kohlhoff, and Patrick Riley.
\newblock Tensor field networks: Rotation- and translation-equivariant neural
  networks for 3d point clouds.
\newblock \emph{ArXiv Preprint}, 2018.

\bibitem[Uy et~al.(2019)Uy, Pham, Hua, Nguyen, and
  Yeung]{uy-scanobjectnn-iccv19}
Mikaela~Angelina Uy, Quang-Hieu Pham, Binh-Son Hua, Duc~Thanh Nguyen, and
  Sai-Kit Yeung.
\newblock Revisiting point cloud classification: A new benchmark dataset and
  classification model on real-world data.
\newblock In \emph{International Conference on Computer Vision (ICCV)}, 2019.

\bibitem[van Steenkiste et~al.(2018)van Steenkiste, Chang, Greff, and
  Schmidhuber]{steenkiste2018relational}
Sjoerd van Steenkiste, Michael Chang, Klaus Greff, and Jürgen Schmidhuber.
\newblock Relational neural expectation maximization: Unsupervised discovery of
  objects and their interactions.
\newblock \emph{International Conference on Learning Representations (ICLR)},
  2018.

\bibitem[Vaswani et~al.(2017)Vaswani, Shazeer, Parmar, Uszkoreit, Jones, Gomez,
  Kaiser, and Polosukhin]{VaswaniSPUJGKP17}
Ashish Vaswani, Noam Shazeer, Niki Parmar, Jakob Uszkoreit, Llion Jones,
  Aidan~N. Gomez, Lukasz Kaiser, and Illia Polosukhin.
\newblock Attention is all you need.
\newblock \emph{Advances in Neural Information Processing Systems (NeurIPS)},
  2017.

\bibitem[Veličković et~al.(2018)Veličković, Cucurull, Casanova, Romero,
  Liò, and Bengio]{VelikovicCCRLB2017}
Petar Veličković, Guillem Cucurull, Arantxa Casanova, Adriana Romero, Pietro
  Liò, and Yoshua Bengio.
\newblock Graph attention networks.
\newblock \emph{International Conference on Learning Representations (ICLR)},
  2018.

\bibitem[Wagstaff et~al.(2019)Wagstaff, Fuchs, Engelcke, Posner, and
  Osborne]{Wagstaff2019}
Edward Wagstaff, Fabian~B. Fuchs, Martin Engelcke, Ingmar Posner, and
  Michael~A. Osborne.
\newblock On the limitations of representing functions on sets.
\newblock \emph{International Conference on Machine Learning (ICML)}, 2019.

\bibitem[Wang et~al.(2017)Wang, Girshick, Gupta, and He]{wang2017nonlocal}
Xiaolong Wang, Ross Girshick, Abhinav Gupta, and Kaiming He.
\newblock Non-local neural networks.
\newblock \emph{IEEE Conference on Computer Vision and Pattern Recognition
  (CVPR)}, 2017.

\bibitem[Wang et~al.(2019)Wang, Sun, Liu, Sarma, Bronstein, and Solomon]{DGCNN}
Yue Wang, Yongbin Sun, Ziwei Liu, Sanjay~E. Sarma, Michael~M. Bronstein, and
  Justin~M. Solomon.
\newblock Dynamic graph cnn for learning on point clouds.
\newblock \emph{ACM Transactions on Graphics (TOG)}, 2019.

\bibitem[Weiler and Cesa(2019)]{e2cnn}
Maurice Weiler and Gabriele Cesa.
\newblock {General E(2)-Equivariant Steerable CNNs}.
\newblock In \emph{Conference on Neural Information Processing Systems
  (NeurIPS)}, 2019.

\bibitem[Weiler et~al.(2018)Weiler, Geiger, Welling, Boomsma, and
  Cohen]{WeilerGWBC18}
Maurice Weiler, Mario Geiger, Max Welling, Wouter Boomsma, and Taco Cohen.
\newblock 3d steerable cnns: Learning rotationally equivariant features in
  volumetric data.
\newblock In \emph{Advances in Neural Information Processing Systems
  (NeurIPS)}, 2018.

\bibitem[Winkels and Cohen(2018)]{WinkelsCohen}
Marysia Winkels and Taco~S. Cohen.
\newblock 3d g-cnns for pulmonary nodule detection.
\newblock \emph{1st Conference on Medical Imaging with Deep Learning (MIDL)},
  2018.

\bibitem[Worrall and Welling(2019)]{Worrall19}
Daniel~E. Worrall and Max Welling.
\newblock Deep scale-spaces: Equivariance over scale.
\newblock In \emph{Advances in Neural Information Processing Systems
  (NeurIPS)}, 2019.

\bibitem[Worrall et~al.(2017)Worrall, Garbin, Turmukhambetov, and
  Brostow]{WorrallGRB2016}
Daniel~E. Worrall, Stephan~J. Garbin, Daniyar Turmukhambetov, and Gabriel~J.
  Brostow.
\newblock Harmonic networks: Deep translation and rotation equivariance.
\newblock \emph{IEEE Conference on Computer Vision and Pattern Recognition
  (CVPR)}, 2017.

\bibitem[Wu et~al.(2015)Wu, Song, Khosla, Yu, Zhang, Tang, and
  Xiao]{modelnet40}
Zhirong Wu, Shuran Song, Aditya Khosla, Fisher Yu, Linguang Zhang, Xiaoou Tang,
  and Jianxiong Xiao.
\newblock 3d shapenets: A deep representation for volumetric shapes.
\newblock \emph{IEEE Conference on Computer Vision and Pattern Recognition
  (CVPR)}, 2015.

\bibitem[Xie et~al.(2018)Xie, Liu, and Tu]{ShapeContextNet}
Saining Xie, Sainan Liu, and Zeyu Chen~Zhuowen Tu.
\newblock Attentional shapecontextnet for point cloud recognition.
\newblock \emph{IEEE Conference on Computer Vision and Pattern Recognition
  (CVPR)}, 2018.

\bibitem[Xu et~al.(2018)Xu, Fan, Xu, Zeng, and Qiao]{SpiderCNN}
Yifan Xu, Tianqi Fan, Mingye Xu, Long Zeng, and Yu~Qiao.
\newblock Spidercnn: Deep learning on point sets with parameterized
  convolutional filters.
\newblock \emph{European Conference on Computer Vision (ECCV)}, 2018.

\bibitem[Yang et~al.(2019)Yang, Zhang, and Ni]{Yang2019}
Jiancheng Yang, Qiang Zhang, and Bingbing Ni.
\newblock Modeling point clouds with self-attention and gumbel subset sampling.
\newblock \emph{IEEE Conference on Computer Vision and Pattern Recognition
  (CVPR)}, 2019.

\bibitem[You et~al.(2019)You, Lou, Liu, Tai, Ma, Lu, and Wang]{You2019}
Yang You, Yujing Lou, Qi~Liu, Yu-Wing Tai, Lizhuang Ma, Cewu Lu, and Weiming
  Wang.
\newblock {Pointwise Rotation-Invariant Network with Adaptive Sampling and 3D
  Spherical Voxel Convolution}.
\newblock In \emph{AAAI Conference on Artificial Intelligence}, 2019.

\bibitem[Zaheer et~al.(2017)Zaheer, Kottur, Ravanbhakhsh, P{\'{o}}czos,
  Salakhutdinov, and Smola]{Zaheer2017}
Manzil Zaheer, Satwik Kottur, Siamak Ravanbhakhsh, Barnab{\'{a}}s P{\'{o}}czos,
  Ruslan Salakhutdinov, and Alexander Smola.
\newblock {Deep Sets}.
\newblock In \emph{Advances in Neural Information Processing Systems
  (NeurIPS)}, 2017.

\bibitem[Zhang et~al.(2019)Zhang, Hua, Rosen, and Yeung]{zhang2019rotation}
Zhiyuan Zhang, Binh-Son Hua, David~W. Rosen, and Sai-Kit Yeung.
\newblock Rotation invariant convolutions for 3d point clouds deep learning.
\newblock In \emph{International Conference on 3D Vision (3DV)}, 2019.

\end{thebibliography}
